# Multi-Objective Path Planning of an Autonomous Mobile Robot using Hybrid PSO-MFB Optimization Algorithm

Fatin H. Ajeil[1], Ibraheem Kasim Ibraheem[1], Mouayad A. Sahib[2]*, Amjad J. Humaidi[3]

[1]Electrical Engineering Department, College of Engineering, University of Baghdad, Baghdad, Iraq.
[2]College of Engineering, University of Information Technology and Communications, Baghdad, Iraq.
[3]Department of Control and Systems Engineering, University of Technology, Baghdad, Iraq.

*Corresponding Author
Email: mouayad.sahib@uoitc.edu.iq

**Abstract:**

The main aim of this paper is to solve a path planning problem for an autonomous mobile robot in static and dynamic environments. The problem is solved by determining the collision-free path that satisfies the chosen criteria for shortest distance and path smoothness. The proposed path planning algorithm mimics the real world by adding the actual size of the mobile robot to that of the obstacles and formulating the problem as a moving point in the free-space. The proposed algorithm consists of three modules. The first module forms an optimized path by conducting a hybridized Particle Swarm Optimization-Modified Frequency Bat (PSO-MFB) algorithm that minimizes distance and follows path smoothness criteria. The second module detects any infeasible points generated by the proposed PSO-MFB Algorithm by a novel Local Search (LS) algorithm integrated with the PSO-MFB algorithm to be converted into feasible solutions. The third module features obstacle detection and avoidance (ODA), which is triggered when the mobile robot detects obstacles within its sensing region, allowing it to avoid collision with obstacles. The simulation results indicate that this method generates an optimal feasible path even in complex dynamic environments and thus overcomes the shortcomings of conventional approaches such as grid methods. Moreover, compared to recent path planning techniques, simulation results show that the proposed PSO-MFB algorithm is highly competitive in terms of path optimality.

***Keywords**: Autonomous Mobile Robot, Robot path planning, particle swarm optimization, bat algorithm, collision avoidance.*

## 1. Introduction

Autonomous navigation of mobile robots cover a wide spectrum of applications, including mining, military, rescuing, space, agriculture, and entertainment [1]. In these applications, successful navigations of mobile robots mainly depend on their intelligence capabilities. Among these capabilities, path planning is the most effective and important intelligent feature. Path planning involves the creation of an optimized collision-free path from one place to another. It can be divided into various categories depending on the nature of the environment: static path planning, where obstacles do not change their position with time, and dynamic path planning where the position and orientation of obstacles change with time. These can be further subdivided according to the knowledge level of the mobile robot into offline and online algorithms. In offline path planning, the mobile robot has a complete knowledge of the environment. Consequently, the path planning algorithm produces a complete path before the robot begins moving. However, in online path planning, information about the environment is obtained from a local sensor attached to the mobile robot, and the mobile robot requires the ability to construct a new path in response to any environment change [2]. This categorization can be further subcategorized according to the target nature, into stationary and dynamic targets. In stationary target, the mobile robot searches for a static point in its workspace, and once it has located this point, it will never move away from it. On the other hand, in dynamic target, the mobile robot must search for a moving target while avoiding obstacles. In the latter case, the mobile robot and its target are both moving [3]. Each of the above scenarios require different path planning algorithms.

Path planning studies began in the late 1960s, and various techniques have been suggested involving cell decomposition [4], roadmap approaches [5], and potential fields [6]. The main drawbacks of these algorithms are inefficiency, because of high computational costs, and inaccuracy, because of the high risk of getting stuck in local minima. Adopting various heuristic methodologies can defeat the impediments of these algorithms, and these include the application of neural systems, genetics, and nature-inspired algorithms [7]. Rapid satisfactory solutions are one of the leading significant focal points of such heuristic methodologies, and these are particularly appropriate for finding solutions to NP-complete problems.

In this paper a new path planning algorithm is developed. The algorithm consists of three main modules: the first module involves point generation achieved using a novel heuristic nature-inspired algorithm, which is a hybridization between Particle Swarm Optimization and Modified Frequency Bat Algorithms (PSO-MFB). The PSO-MFB generates and select the points that satisfy the multi-objective measure proposed in this work, which is a combination of shortest path and path smoothness. Then the proposed algorithm is integrated with a second module which converts infeasible solutions into feasible ones. In the third module an avoidance algorithm is employed to avoid obstacles.

The current paper is structured as follows. First, section 2 highlights several research methodologies, then section 3 presents the problem statement, preliminaries, and the performance criteria considered in this work. Swarm based optimization is introduced in section 4, while in section 5, the methodologies proposed for mobile robot path planning in this work are introduced. In section 6, a set of simulation results are presented to demonstrate the effectiveness of the proposed methodology as compared with previous works. Section 7 present a discussion of the obtained results. Finally. Conclusions and recommendations are presented in section 8.

## 2. Related Works

Numerous approaches have been used to solve single/multi-objective path planning problems for mobile robots, such as swarm/nature-inspired algorithms, neural networks, and fuzzy logic. The first group includes several previous studies that have exploited examples of natural swarm behaviours. The works in [8] and [9] utilized the standard Ant Colony optimization (ACO) to solve path planning problems for complex environments. An improved version of ACO (IACO) has been proposed in [10] to obtain faster convergence speed and to avoid trapping into local minimum. Compared to other algorithms, the IACO produced optimum path, however, it takes longer time to converge [10].

Various works have also adopted heuristic methods and employed these to solve different aspects of path planning methods such as Bat algorithm (BA) [11], Particle Swarm Optimization [12], Cuckoo search (CS) algorithms [13], Bacterial Foraging optimization [14], Artificial Immune Systems [15], and the Whale Optimization Algorithm (WOA), implemented in a static environment to satisfy requirements for the shortest and smoothest path [16]. GA and its modified versions are frequently implemented to find the shortest path for mobile robot path planning in different environments [17], while path planning using neural networks was developed in [18]. Integrating a path planning algorithm with the motion controllers of mobile robots was achieved in [19–22], where several different motion control strategies were employed, including fuzzy logic controls, adaptive neuro-fuzzy inference systems, and model predictive controls. The Wind Driven Optimization (WDO) and Invasive Weed Optimization (IWO) algorithms were used to tune the parameters of the fuzzy logic controller and adaptive neuro-fuzzy inference systems in [20], [21], respectively, while ACO and PSO were used in the tuning of the fuzzy logic controller presented by [23]. The works in [24–26] incorporated two-level navigation algorithms, where the higher level was mainly concerned with path planning and guidance for the mobile robot, while the motion control directing the mobile robot in its configuration space was included in the lowest level. Mobile robot energy consumption is an important issue directly related to smooth trajectory planning. Minimum-energy cornering trajectory planning algorithm with self-rotation and energy constrained objective measures were developed in [27].

Hybridization of meta-heuristic algorithms were also employed to improve robot path planning algorithms. The objective of hybridizing two meta-heuristic algorithms is to combine the advantages of each algorithm to form an improved one. Such hybridized algorithms used in robot path planning include genetic algorithm and particle swarm optimization (GA-PSO) [28], Multi-Objective Bare Bones Particle Swarm Optimization with Differential Evolution (MOBBPSO) [29], cuckoo search (CS) and bat algorithm (BA) [30]

One of the drawbacks in the studies mentioned above is that the mobile robot was treated as a simple particle. While some of these algorithms were oriented toward finding the shortest path avoiding static obstacles. Other studies focused on the avoidance of dynamic obstacles while achieving the shortest distance without considering the smoothness of the path. Moreover, despite the ease of implementation of the grid-based methods used by some of the above researches, these have several disadvantages such as the imprecise representation of the obstacle, where if the obstacle occupies only a small area of the cell, the entire cell is nevertheless reserved for that obstacle. This leads to the waste of a space and less flexibility in dynamic environments.

The main contribution of this paper is to develop a new path planning algorithm which consists of three main modules:

- The first module involves point generation, achieved using a novel heuristic nature-inspired algorithm, which is a hybridization between Particle Swarm Optimization and Modified Frequency Bat Algorithms, thus a PSO-MFB Algorithm. This hybridized algorithm generates and selects the points that satisfy the multi-objective measure proposed in this work, which is a combination of shortest path and path smoothness.

- In the second module the PSO-MFB algorithm is integrated with a second module in which a local search technique converts infeasible solutions into feasible ones.
- In addition, to avoid obstacles, twelve sensors are deployed around the mobile robot to sense obstacles, and once they are detected, an avoidance algorithm is triggered to avoid obstacles; this is a function of the third module.

## 3. Problem Statement and Preliminaries

Assume a mobile robot at a start position (SP) that is required to reach a goal position (GP) in a given workspace. Several static and dynamic obstacles are assumed to exist in the mobile robot workspace. The objective of a path planning problem is to find an optimum or near-optimum path (safest, shortest, and smoothest) without colliding with a problem, some of the assumptions made in this paper should be made explicit.

**Assumption 1:** The obstacles are represented as circular shapes.

**Assumption 2:** The mobile robot is a physical body; thus, to take into account the actual size of the mobile robot, the obstacles are expanded by the radius of mobile robot ($r_{MR}$), so that the mobile robot can be considered as a point, as shown in Fig. 1.

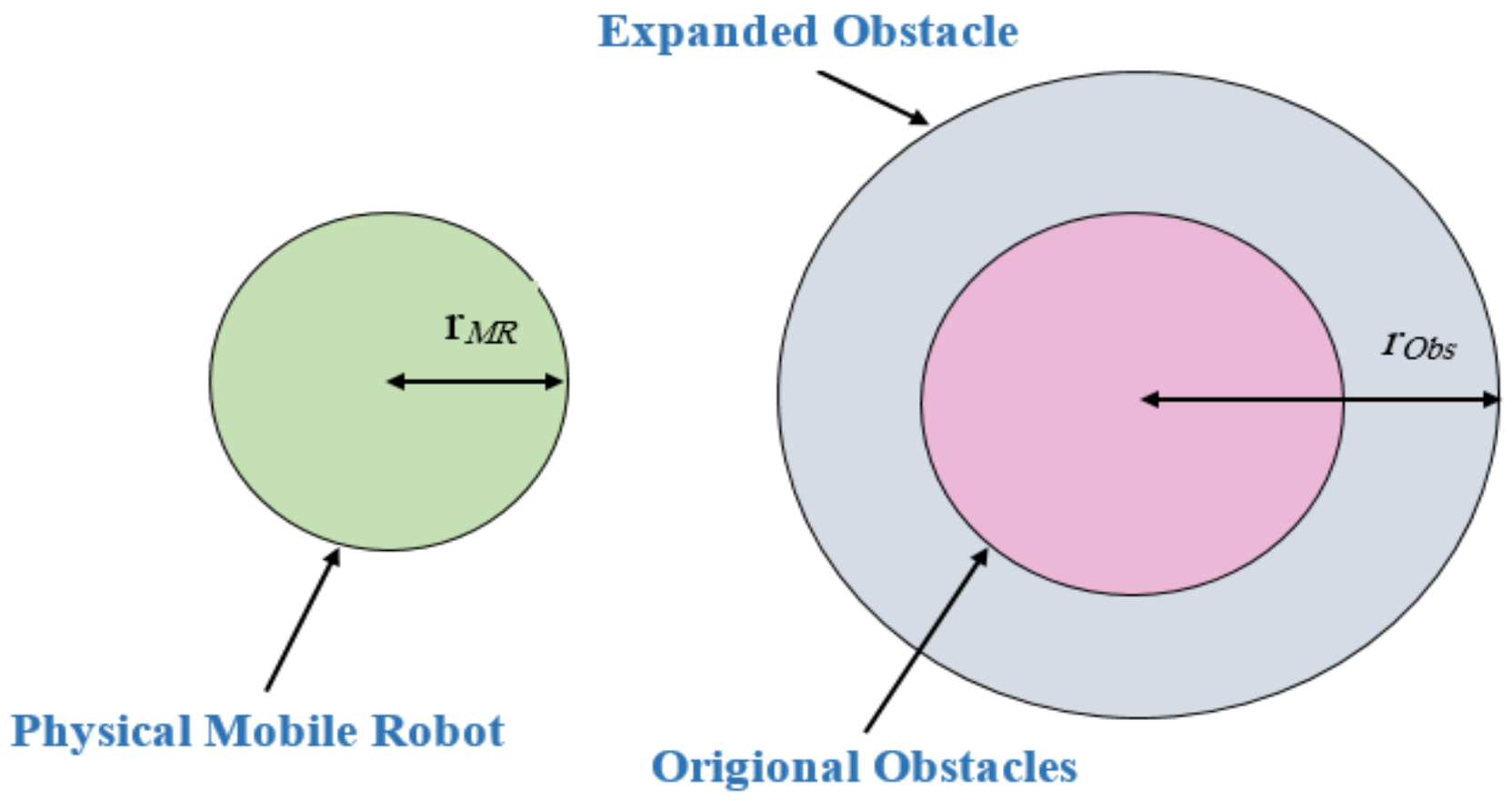

Fig. 1. Expanding obstacles size corresponding to mobile robot size.

**Assumption 3:** There are no kinematic constraints which affect the motion of the mobile robot. The only effective source is the motion of the obstacles.

**Assumption 4:** The mobile robot motion is omnidirectional, and it can move in any direction at any time.

### *A. Performance Criteria*

#### *1) Shortest Distance*

In path planning field, "*Shortest Distance*" means minimising the path length between the start and goal points. At any iteration, the point $wp_j(t)$ is selected as the best one if it has the shortest distance to the goal point $wp(N)$ such that

$$f_1(x, y) = d(wp_j(t), wp(N)) \tag{1}$$

is minimum, where $d(.,.)$ is the Euclidean distance. The Shortest Path Length (SPL) is the sum of all the distances between mid-points $(wp_j(2) \cdots wp_j(N-1))$ generated by the path planning algorithm between the start point (*SP*) $wp(1)$ and the Goal Point (*GP*) $wp(N)$, given by:

$$SPL = \sum_{t=1}^{N-1} d\left(wp_j(t), wp_j(t+1)\right) = \sum_{t=1}^{N-1} d_t \tag{2}$$

where, *j* is the index of the best solution generated by the swarm optimization based path planning algorithm.

$$d_t = \sqrt{(x_{wp_j(t+1)} - x_{wp_j(t)})^2 + (y_{wp_j(t+1)} - y_{wp_j(t)})^2}$$

*2) Path Smoothness*

This involves minimising the difference of the angles between the straight lines (goal-current points and suggested points-current point), as shown in Fig. 2 and given by:

$$f_2(x, y) = \sum_{t=1}^{N-1} \left|\theta_{(wp(t),wp(t+1))} - \theta_{(wp(t),wp(N))}\right| \tag{3}$$

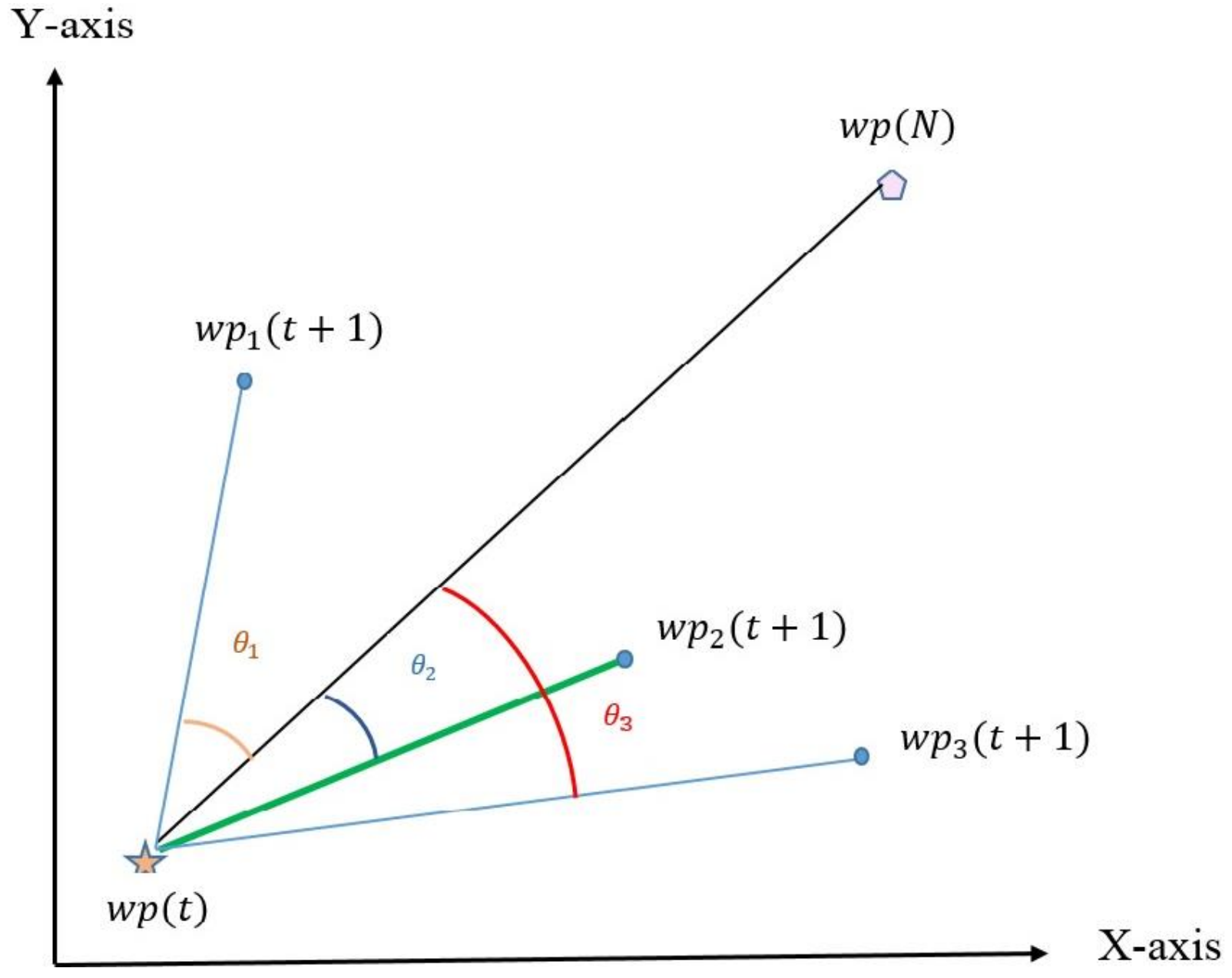

Fig. 2. Path smoothness: Summing Angles errors.

where,

$$\theta_{(wp_{(t)},wp_{(t+1)})} = tan^{-1}\frac{y_{wp_j\ (t+1)} - y_{wp\ (t)}}{x_{wp_j\ (t+1)} - x_{wp\ (t)}}\ ,\ \theta_{(wp(t),wp(N))} = tan^{-1}\frac{y_{wp\ (N)} - y_{wp\ (t)}}{x_{wp\ (N)} - x_{wp(t)}}$$

From Fig. 2, $\theta_1$ , $\theta_2$ , and $\theta_3$ are the angles between line segment $(wp(t), wp(N))$ and $(wp(t),\ wp_1(t+1))$,( $wp(t),\ wp_2(t+1)$),( $wp(t),\ wp_3(t+1)$) respectively. It's obvious that the $\theta_2$ has minimum angle among the competing points ($wp_1$,$wp_2$,$wp_3$).

The overall multi-objective optimization is the weighted sum of the above two objectives:

$$f(x, y) = w_1\ f_1(x, y) + w_2 f_2(x, y) \tag{4}$$

where $w_1$ and $w_2$ are degrees of importance of the two objectives. Their values must satisfy the following condition:

$$w_1 + w_2 = 1 \tag{5}$$

The overall fitness function is given by:

$$fitness = \frac{1}{f(x, y) + \mathcal{E}} \tag{6}$$

where $\mathcal{E}$ is a small number (e.g., $\mathcal{E} = 0.001$) used to prevent division by zero case. The process of selecting the best solution among competing feasible options in each iteration depends on the balance between the two performance objectives declared in (1) and (3) for all available solutions. In Fig. 3, the best point among six competing points is point $wp_3$ for the *t* iteration, and point $wp_5$ in the (*t*+1) iteration, while in the (*t*+2) iteration, points $wp_2$ and $wp_3$ offer shorter distances but larger difference angles, in contrast to point $wp_1$ which provides a balance between the two criteria; thus, point $wp_1$ is selected. This process is continued until reaching *GP*.

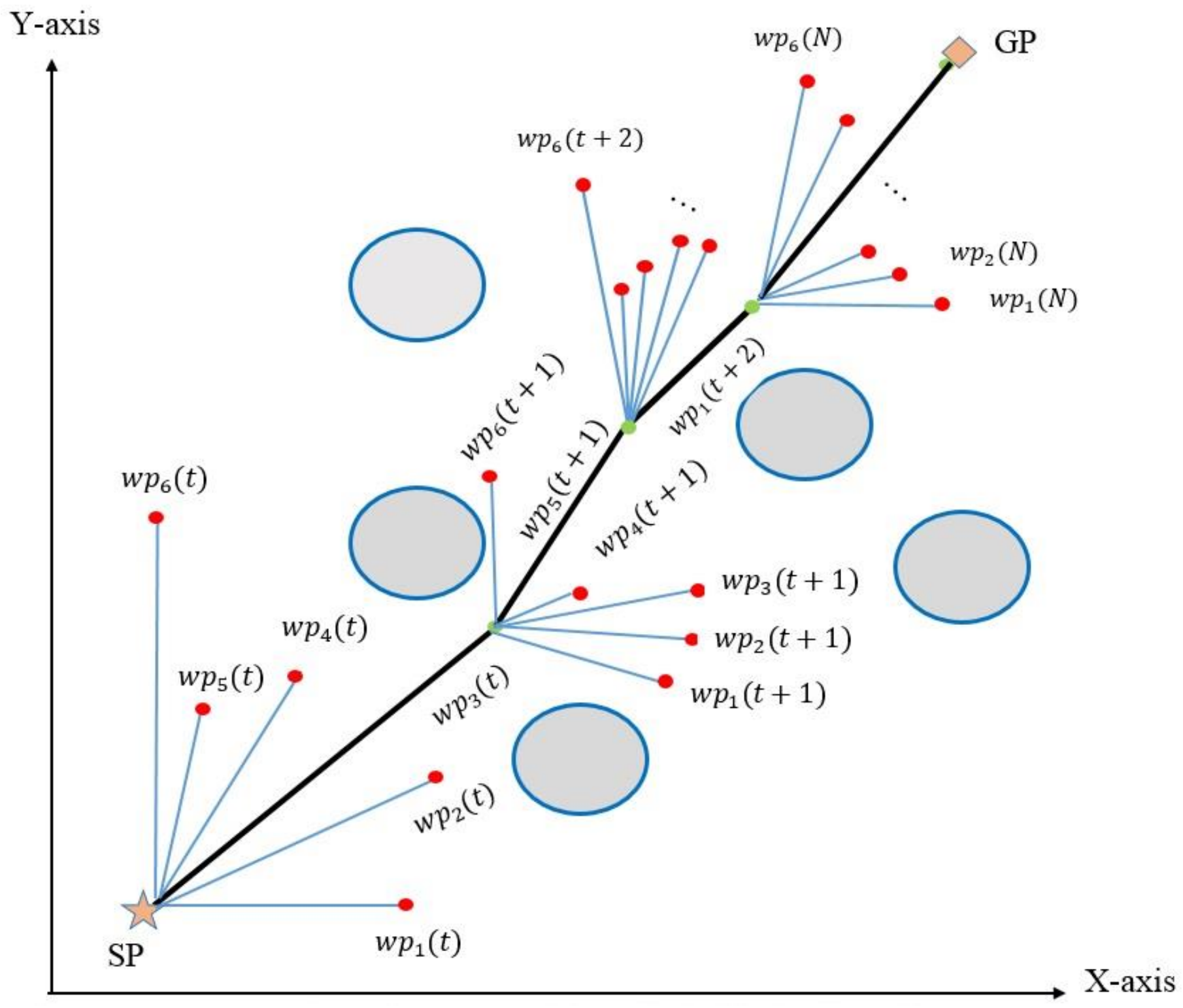

Fig. 3. Mid-points selection for multi-objective path planning.

*B. Obstacles Movement*

In this case, the obstacle changes its location at each time step continuously. The movement of the dynamic obstacles in this work is considered to be one of the following types:

*1) Linear movement*

In this case, the obstacles move in a straight line with specific velocity ($v_{obs}$) and direction ($\varphi_{obs}$) according to the following relationship:

$$x_{obs_New} = x_{obs} + v_{obs} \times cos\, \varphi_{obs} \tag{7}$$

$$y_{obs_New} = y_{obs} + v_{obs} \times sin\, \varphi_{obs} \tag{8}$$

where $\varphi_{obs}$ is the slope of the linear motion.

*2) Circular trajectory*

The obstacles move along a circular path given by the centre of the circle $(x_c, y_c)$ and the radius $(r_c)$. Thus, the new position of the obstacle is given by:

$$x_{obs} = x_c + r_c \times \cos \partial \tag{9}$$

$$y_{obs} = y_c + r_c \times \sin \partial \tag{10}$$

The range of $\partial$ represents the portion of circular arc for a complete circle ($0 < \partial < 2\pi$).

## 4. Swarm-based Optimization

Swarm Intelligence (SI) is an artificial collection of simple agents based on nature-inspired behaviours that can be successfully applied to optimization problems in a variety of applications. The search process of such optimization algorithms continues to find new solutions until a stopping condition is satisfied (either the optimal solution is found, or a maximum number of iterations is reached). These SI behaviours can be used to solve a variety of problems, and thus there are several SI-based algorithms. Two such algorithms are used in this paper.

*A. Particle Swarm Optimization (PSO) Algorithm*

This is a population-based heuristic strategy for optimization problems developed by J. Kennedy and R. C. Eberhard in 1995 [31], stimulated by the social conduct of schooling fish and flocking birds. It consists of a swarm of particles, and each particle in PSO has a position $x_i$ and velocity $v_i$. The position represents a solution suggested by the particle, while the velocity is the rate of change to the next position with respect to the current position. These two values (position and velocity) are randomly initialized, and the solution construction of PSO algorithm includes two phases:

- *Velocity Update of the Particle*

$$v_i(t+1) = v_i(t) + c_1 r_1 (pbest_i - x_i) + c_2 r_2 (gbest - x_i) \tag{11}$$

- *Position Update of the Particle*

$$x_i(t+1) = x_i(t) + v_i(t+1) \tag{12}$$

From (11), the velocity of the particle $i$ is affected by three main components: the particle's old velocity $v_i(t)$; a linear attraction toward the personal best position ever found $(pbest_i - x_i)$, scaled by the weight $c_1$and a random number $r_1 \in [0, 1]$; and the final component is a linear attraction towards the global best position found by any particle in the swarm $(gbest - x_i)$, scaled by weight $c_2$ and a random number $r_2 \in [0, 1]$. The particle's position for the $t+1$ iteration is updated according to (12).

*B. Modified Frequency Bat (MFB) Algorithm*

The Bat Algorithm (BA) is a bio-inspired algorithm developed by Yang in 2010 [32]. It is based on the echolocation or bio sonar characteristics of micro bats. Echolocation is an important feature of bat behaviour: the bats emit sound pulses and listen to the echoes bouncing back from obstacles while flying. By utilising the time difference between its ears, the loudness of the response, and the delay time, a bat can thus figure out the velocity, shape, and size of prey and obstacles. A bat also has the ability to change the way its sonar works. If it sends sound pulses at a high rate, it can fly for less time while obtaining thorough details about its surroundings.

*1) The Movement of Artificial Bats*

The updating process of bat positioning is as follows:

$$f_i = f_{min} + (f_{max} - f_{min}) * \beta_i \tag{13}$$

$$v_i(t+1) = v_i(t) + (z_i(t+1) - z^*)f_i \tag{14}$$

$$z_i(t+1) = z_i(t) + v_i(t+1) \tag{15}$$

where $\beta_i = t * \mathrm{e}^{(-\rho * r)}$, $t$ is iteration number, $r$ is a random number [0,1], the value of $\rho$ is an application dependent, for path planning problem the value of $\rho$ is chosen to be (0.01). This means that $\beta$ will increase with time for each bat generating low frequencies at the early stages of the search process. These frequencies increase with time to improve the global search performance, where $z^*$ is the present global best location solution, found after making a comparison between all the solutions among all $m$ bats. For the local search stage, once a solution is selected among the current best solutions, a new solution is generated for each bat locally using random walk:

$$z_{new} = z_{old} + \sigma \epsilon A(t) \tag{16}$$

where $\epsilon$ is a random number within $[-1, 1]$. $A(t)$ is the average loudness of all the bats at time step $t$, and σ is a scaling parameter included to control the step size.

*2)* Loudness and Pulse Emission

The loudness, $A_i$, and the rate of pulse emission, $r_i$, must be updated as the iterations proceed. The loudness usually decreases once a bat has found its prey, whereas the rate of pulse emission increases according to the following equations:

$$A_i\,(t+1) = \alpha A_i\,(t) \tag{17}$$

$$r_i(t+1) = r_i(0)[1 - \exp(-\gamma t)] \tag{18}$$

where $0 < \alpha < 1$ and $\gamma > 0$ are design parameters.

## 5. Proposed Method

This section describes the proposed path planning algorithm for a mobile robot with omnidirectional motion based on hybridized swarm optimization integrated with Local Search and obstacle avoidance techniques.

*A. Proposed Hybrid PSO-MFB Algorithm*

To increase the overall performance, the advantage features of two or more optimization algorithms are combined to produce a hybridised optimization algorithm. In this paper, a hybridisation between PSO and MFB algorithms is proposed. The variations of loudness, $A_i$, and pulse emission rates, $r_i$, also provide an auto zooming capability for the optimization algorithm. Finding the optimum values of the MFB algorithm parameters ($\alpha$, $\gamma$) is handled by the PSO algorithm. However, such parameter settings may be problem-dependent and thus tricky to define. In addition, the use of time-varying parameters during such iterations may be advantageous. The proper control of such parameters can thus be important, and consequently, variations of the parameters $\alpha$ and $\gamma$ (hence the loudness $A_i$ and the pulse rate $r_i$) within a suitable range have been adapted by the PSO algorithm to find a balance between exploration and exploitation in the MFB algorithm. The pseudo code for the proposed Hybrid PSO-MFB algorithm is shown in **Algorithm 1** and the overall procedure of the proposed Hybrid PSO-MFB algorithm is shown in Fig. 4. The solution of the PSO in the proposed algorithm (particle size) is a vector of dimension 2, where $x(1,1)$ represents the value of $\alpha$ while $x(1,2)$ is the value of $\gamma$.

### *B. Proposed Local Search (LS) Technique*

The solution is considered infeasible if the next point generated by the hybrid PSO-MFB algorithm lies within an area occupied by an obstacle. Another infeasible solution is where the next point $wp_c(t+1)$ and the previous point $wp(t)$ form a line segment passing through an obstacle.

**Algorithm 1:** Pseudo code for Proposed Hybrid PSO-MFB

Initialize PSO and MFB parameters: population size of $n$ particles, $r_1$, $r_2$, $c_1$, $c_2$, population size of $m$ bats, frequencies $f_i$, pulse rates $r_i$ and the loudness $A_i$;<br>
Randomly generate an initial solutions, $x_1 = [\alpha_1, \gamma_1]$, $x_2 = [\alpha_2, \gamma_2]$, …, $x_n = [\alpha_n, \gamma_n]$;<br>
**for** $i$ = 1: $n$<br>
    Call MFB algorithm, (13)-(18);<br>
**end for**<br>
Choose the best bat that achieves the best fitness defined in (6), e.g., $z_{k\lambda}$, $\lambda$=1: $m$;<br>
Store index ($k$);<br>
$Gbest = x_k = [\alpha_k, \gamma_k]$;<br>
**If** stopping criteria not satisfied **then**<br>
    Update velocity and position of particles according to (11)-(12);<br>
    Go to 3;<br>
**Else**<br>
    obtain results;<br>
**end if**

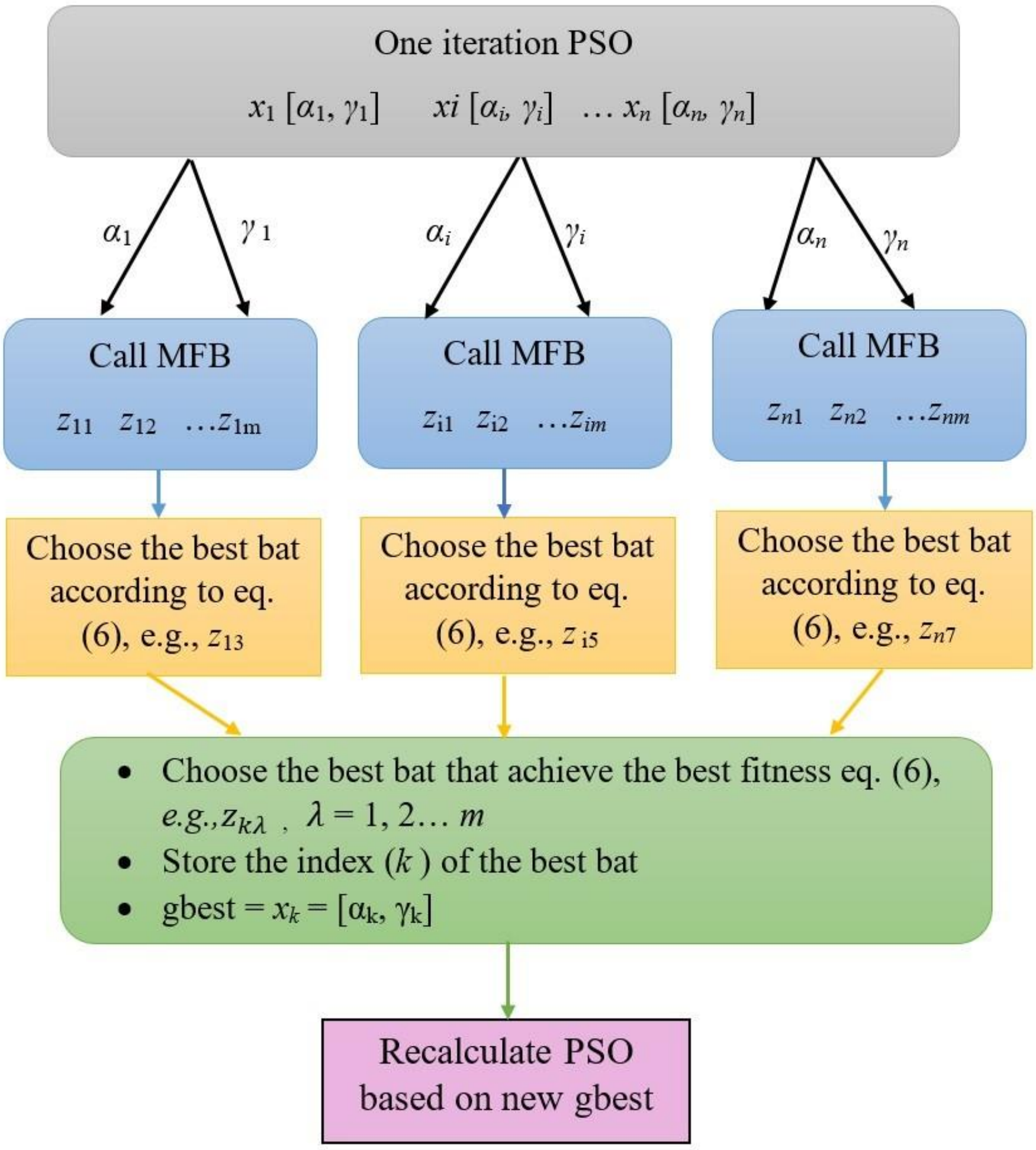

Fig. 4. Proposed Hybrid PSO-MFB optimization algorithm.

The LS technique converts these infeasible solutions into feasible ones. These two situations are explained in the following section with the aid of graphical and mathematical illustrations of the proposed solutions.

*1) The points lie inside the obstacle*

This situation is shown in Fig. 5 (a). It is checked by computing the Euclidean distance $d(wp_c(t+1), P_{obs})$ using (1) between the candidate point, $wp_c(t+1)$, and the centre of the obstacle $P_{obs} = (x_{obs}, y_{obs})$. If $d(wp_c(t+1), P_{obs})$or simply $d$, is less than the obstacle's radius, $r_{Obs}$, then it is considered an infeasible candidate solution:

$$d(wp_c(t+1), P_{obs}) < r_{Obs} \tag{19}$$

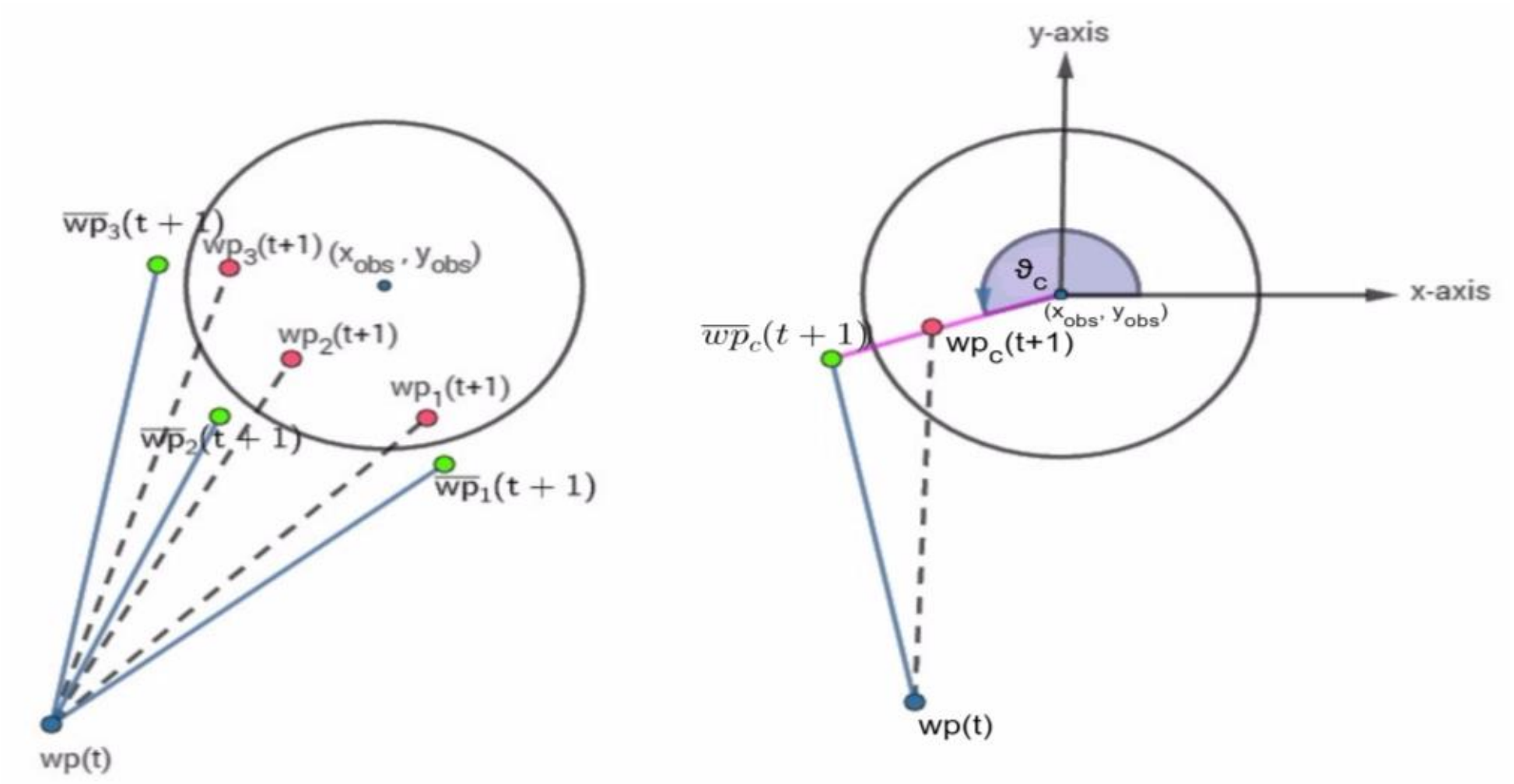

Fig. 5. Infeasible path, (a) the point lies inside the obstacle, (b) the proposed solution.

This case is resolved by ousting these candidate solutions outside the area occupied by the obstacle according to the following suggested rules; see Fig. 5(b):

$$x_{\overline{wp}_c(t+1)} = x_{wp_c(t+1)} + (r_{Obs} + ds - d) \times \cos\vartheta_c \tag{20}$$

$$y_{\overline{wp}_c(t+1)} = y_{wp_c(t+1)} + (r_{Obs} + ds - d) \times \sin\vartheta_c \tag{21}$$

where $ds$ refers to the minimum safety distance, $\vartheta_c$ is the angle between obstacle centre, and *c-th* candidate point $wp_c(t+1)$. Therefore, the red points in Fig. 5 represent the candidate solutions generated by the hybrid PSO-MFB Algorithm, while the green ones are the updated ones according to (20) and (21), where $\overline{wp}_c(t+1)$ =($x_{\overline{wp}_c(t+1)}$, $y_{\overline{wp}_c(t+1)}$).

*2) The line segment passes through the obstacle*

The second situation is shown in Fig. 6 (a); here, a line segment that connects two consecutive points, $wp(t)$ and $wp_c(t+1)$, passes through the region occupied by the obstacle. This situation can be resolved using the following procedure:

1- Find the equation of the line segment that connects any two consecutive points as shown in Fig.6 (b).

$$y = mx + q \tag{22}$$

Where: $m$ is slope of line segment, $q$ is dc value

Substitute $y$ as given by (22) into the circle equation that describes the obstacle circumference:

$$(x - x_{obs})^2 + (y - y_{obs})^2 = (r_{Obs})^2 \tag{23}$$

2- Solve for $x$ to find whether the line intersects with the obstacle. The resulting equation will be quadratic and in terms of only x, and its solution is given as:

$$x_{1,2} = \frac{-b \pm \sqrt{b^2 - 4ac}}{2a} \tag{24}$$

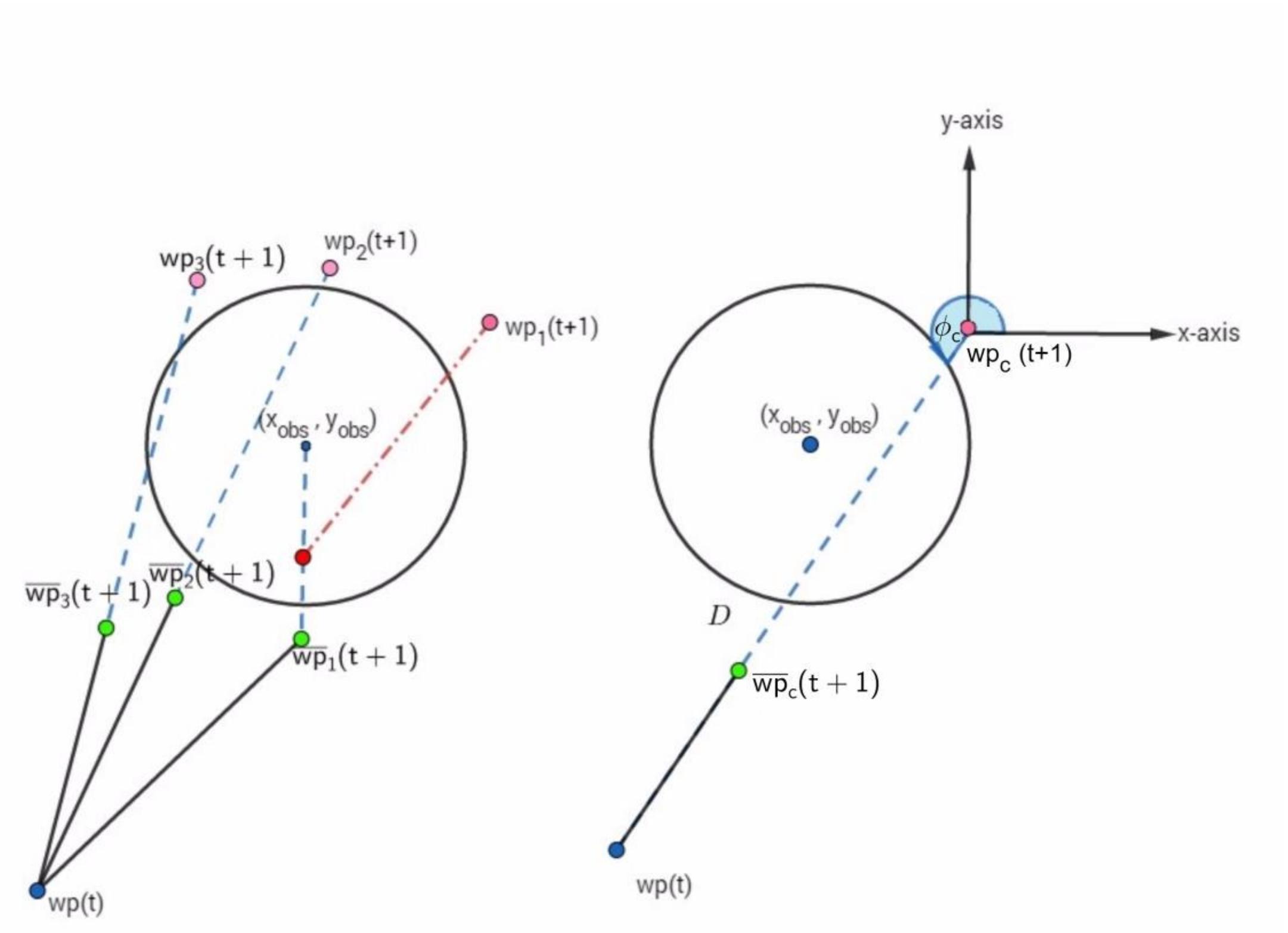

(a) (b)

Fig. 6. Infeasible Path: (a) the line segment passes through the obstacle, (b) the proposed solution.

The exact behaviour depends on $b^2 - 4ac$ and is determined by the three possible solutions:

- if $b^2 - 4ac < 0$, then $x_{1,2}$ are complex. Here, the path segment does not intersect with the obstacle and the path is a *feasible solution*.
- if $b^2 - 4ac = 0$, then $x_{1,2}$ has a single solution. The path segment is tangential to the obstacle and the path is considered to be an *infeasible solution*.
- Finally, if $b^2 - 4ac > 0$, then $x_{1,2}$ are real, the path segment intersects with the obstacle, and the path is an *infeasible solution*.

The solutions are updated by ousting the next candidate solutions outside the region of the obstacle according to the following rules (green points in Fig.6 (b)):

$$x_{\overline{wp}_c(t+1)} = x_{wp_c(t+1)} + (\delta * D)\cos \emptyset_c \quad (25)$$
$$y_{\overline{wp}_c(t+1)} = y_{wp_c(t+1)} + (\delta * D)\sin \emptyset_c \quad (26)$$

$\delta$ is chosen to be 0.6, $D$ is the distance between candidate way point ($wp_c(t+1)$) and pervious way point ($wp(t)$)

*C. Obstacle Detection and Avoidance (ODA)*

When a moving obstacle gets closer to the mobile robot while the latter follows the feasible path generated by the proposed hybrid PSO-MFB algorithm, or the mobile robot itself gets closer to a static obstacle, it must instantly react to avoid this obstacle, or a collision will occur. In this section, the sensors deployment and a proposed method for sensing obstacles to achieve obstacle detection is thus presented, and a proposed method for avoiding these obstacles called the gap vector method is discussed.

*1) Obstacle detection (OD) Procedure*

Sensing is accomplished by attaching twelve virtual sensors around the mobile robot. These are separated equally, with each sensor covering an angle range of 30° and having a certain value of Sensing Range (*SR*), as shown in Fig. 7. As assumed in section 3, the obstacles are expanded by the radius of the mobile robot ($r_{MR}$), so that it can be considered as a point. In this critical worst case, the mobile robot will touch the obstacle, however, for practical considerations an additional distance is added to avoid such case. This distance is provided by the robot sensors and termed as sensing range (*SR*) which is set to 0.8 m in the design.

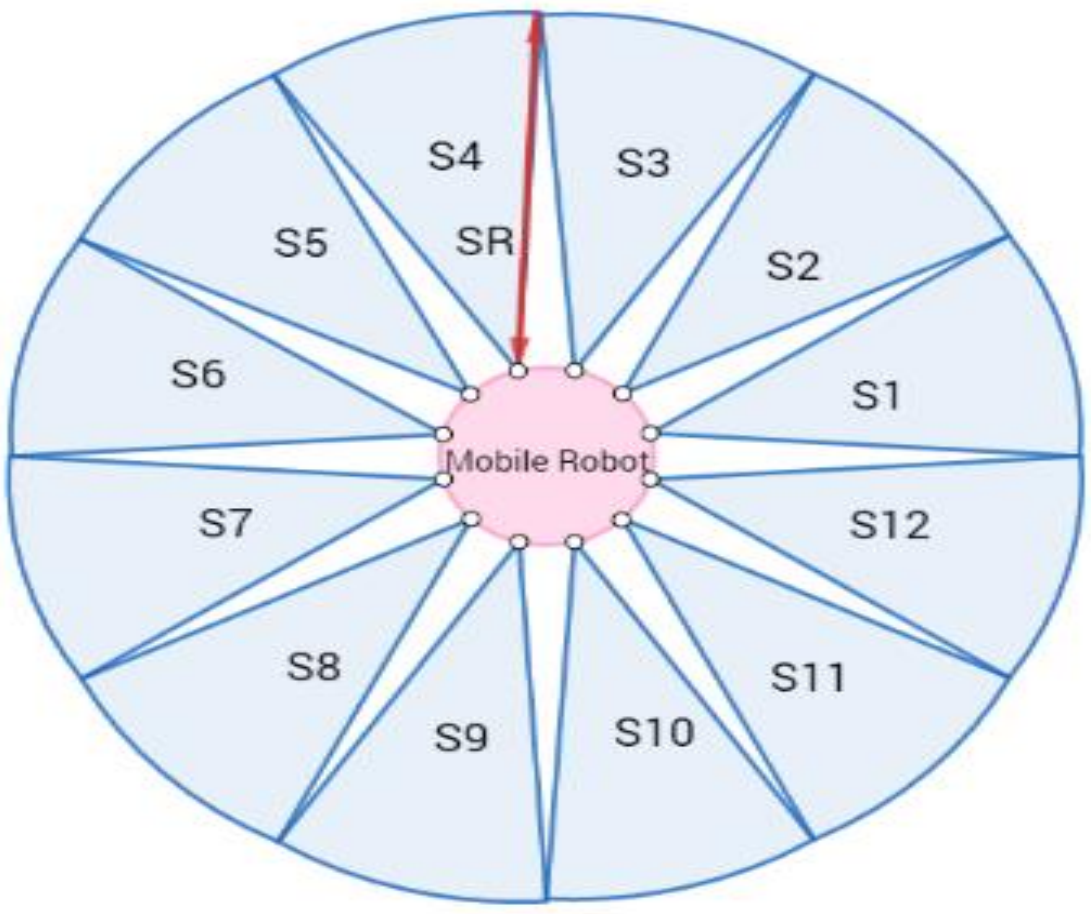

Fig.7. Sensors deployment of the mobile robot.

The obstacles are detected using a *Sensory Vector* (*Vs*) with length equal to the number of deployed sensors:

$$Vs = [a(1) \quad \dots \quad a(i) \quad \dots \quad a(12)]$$

where $a(i)$, $i \in \{1,2,..,12\}$ are variables with binary values, and *Vs* reflects the status of an obstacle extant in an angle range $Si$, $i \in \{1,2,..,12\}$. For example, with $a(1) = a(2) = a(7)$ = logic "1", this indicates that obstacles are detected inside SR and in the angle range $S1$, $S2$, and $S7$ respectively, while a logic "0" in a certain $a(i)$s of *Vs* represents a free space in the corresponding angle ranges $Si$s. To find *Vs*, for each obstacle located inside *SR* in a certain angle range, say $Si$, draw the tangent lines to the expanded obstacle (red circle in Fig. 8) to intersect at the mobile robot A, $MR_{Pos} = (x_{MR}, y_{MR})$. See Fig.8.

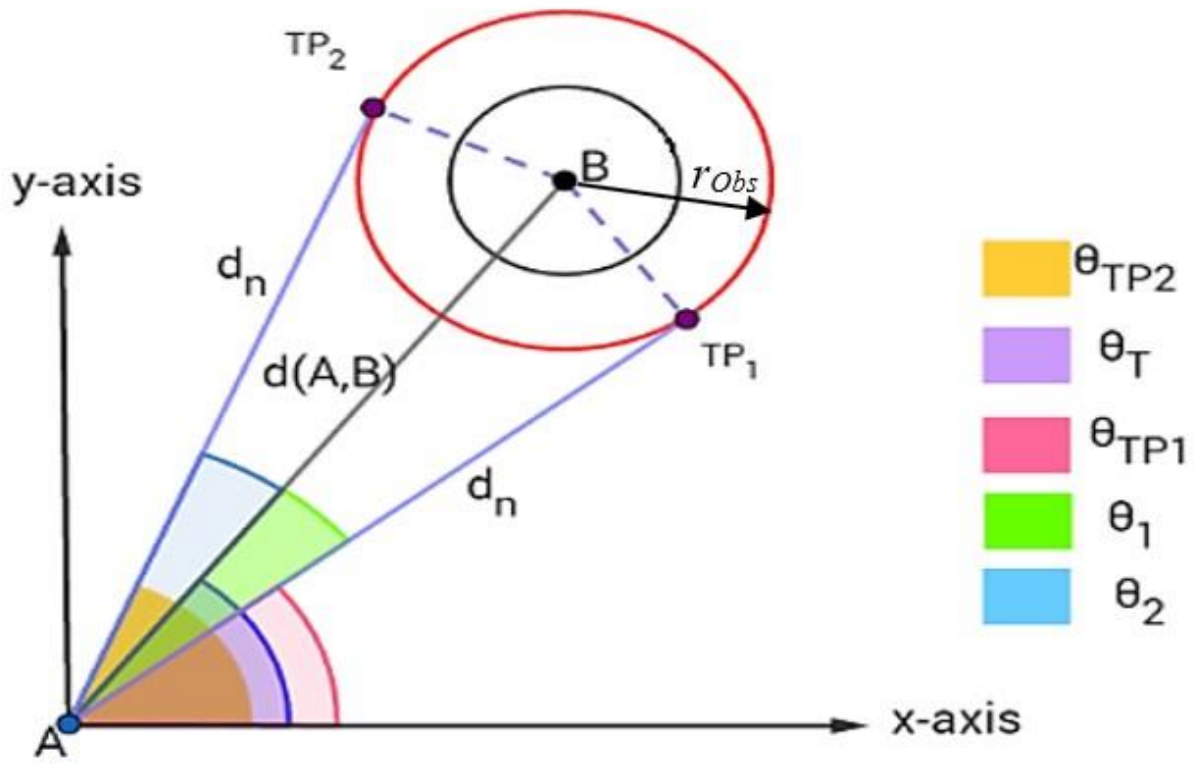

Fig.8. Angle calculation between the mobile robot and the obstacle

Suppose that the distance between the mobile robot $A$ with position $MR_{Pos} = (x_{MR}, y_{MR})$ and the centre of the obstacle $B$ with position $obsB_{Pos} = (x_{obsB}, y_{obsB})$ is Euclidean distance $d(MR_{Pos}, obsB_{Pos})$, or simply $d$ as given by (1), compute angles $\theta_1$ and $\theta_2$ between the hypotenuse d and the tangent lines of the obstacle B extant in a certain *Si*. Given $MR_{Pos} = (x_{MR}, y_{MR})$ and $obsB_{Pos} = (x_{obsB}, y_{obsB})$, $\theta_T$, which describes the angle between the mobile robot A and the obstacle B, can be easily found. From the basic geometry of triangles and since $r_{Obs} \perp d_n$ , property of tangents, then $d^2 = d_n^2 + r_{Obs}^2$ , Pythagoras theorem, thus:

$$\theta_1 = \theta_2 = \sin^{-1}\frac{r_{Obs}}{d} \quad (27)$$

$$\theta_{Tp1} = \theta_T - \theta_1 \quad (28)$$

$$\theta_{Tp2} = \theta_T + \theta_1 \quad (29)$$

Based on the above analysis, *Vs* is found by setting the values of *a*(*i*)'s, $i \in \{1,2,..,12\}$ in the *Vs* to logic "1" if the corresponding angle range *Si* is occupied by static and/or dynamic obstacles. This is realized when the angle difference $\theta_{Tp2} - \theta_{Tp1}$for each obstacle lies in the angle range *Si*'s. An example for the obstacle detection procedure using mobile robot path planning is depicted in Fig. 9; in this case,

*Vs* = [1 1 0 0 0 0 1 1 1 0 0 0]

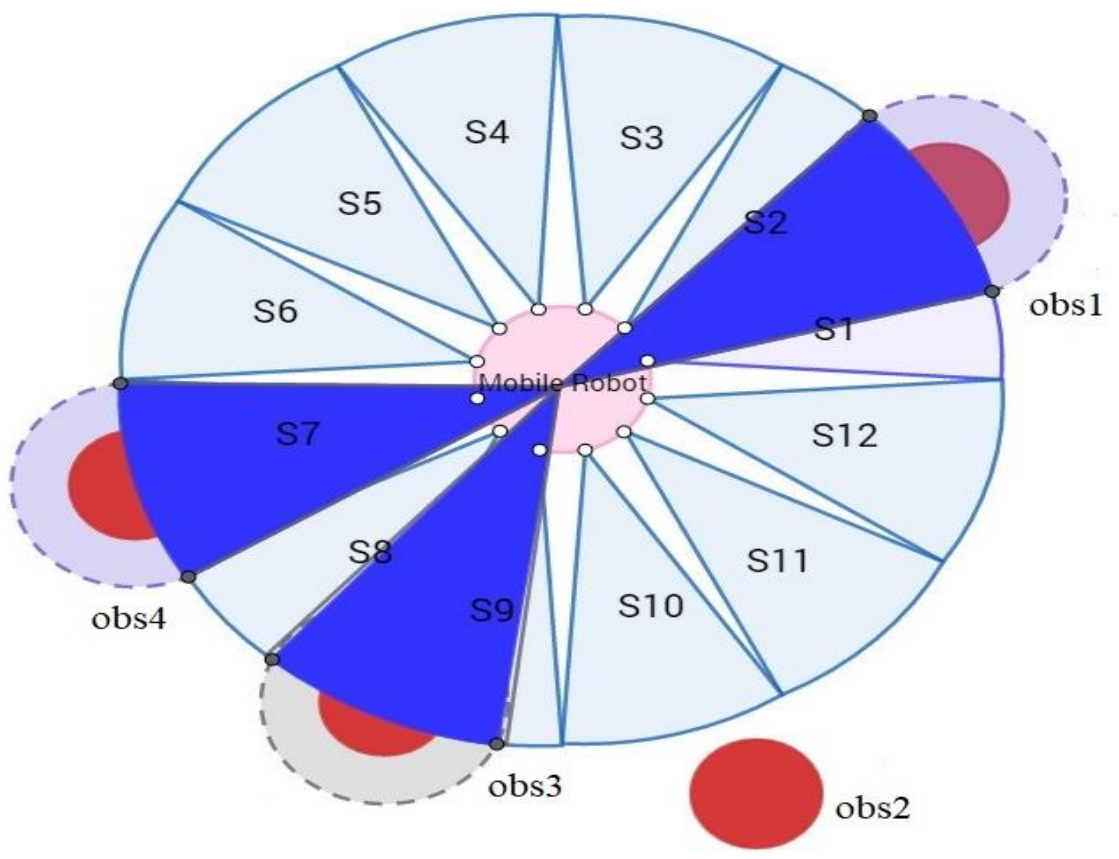

Fig. 9. Obstacle detection of the mobile robot.

The obstacles $obs1$, $obs4$, and $obs3$ lie inside $SR$, and $obs1$ has an angle difference of $\theta_{Tp2} - \theta_{Tp1}$ that lies in the angle ranges $S1$ and $S2$, $obs3$ has its angle difference which lies in the angle ranges $S8$ and $S9$. While obs4 has its own angle difference inside $S7$ only.

*2) Obstacle Avoidance (OA) Algorithm*

Obstacles avoidance is achieved by using a gap vector ($Vg$) concept, which is a binary vector where logic "1" represents an occupancy gap and logic "0" represents a free gap. The length of the gap vector $Vg$ is equal to the length of $Vs$. The mobile robot chooses the gap that gives the shortest path moving towards $GP$. This $Vg$ can be derived from the sensing vector $Vs$ as follows: each consecutive zero in $Vs$ represents a free gap (i.e., logic 0 in $Vg$); otherwise it is an occupied gap (i.e., logic 1 in $Vg$). The above procedure yields:

$Vs$ = [1 1 0 0 0 0 1 1 1 0 0 0]

$Vg$ = [1 1 0 0 0 1 1 1 1 0 0 1]

After constructing $Vg$, several free gaps (permissible suggested mobile robot positions) are produced (see Fig. 10). The angle of each available free gap $\psi_i$ is simply $\psi_\mu = \mu * 30$, where $\mu$ is the index of the "0" in $Vg$. The next step is to determine the next position for the mobile robot (best free gap $gi$ in $Vg$), through which the mobile robot will evade the obstacles and continue moving toward GP using **Algorithm 1**. **Algorithm 2** describes the OA steps with details.

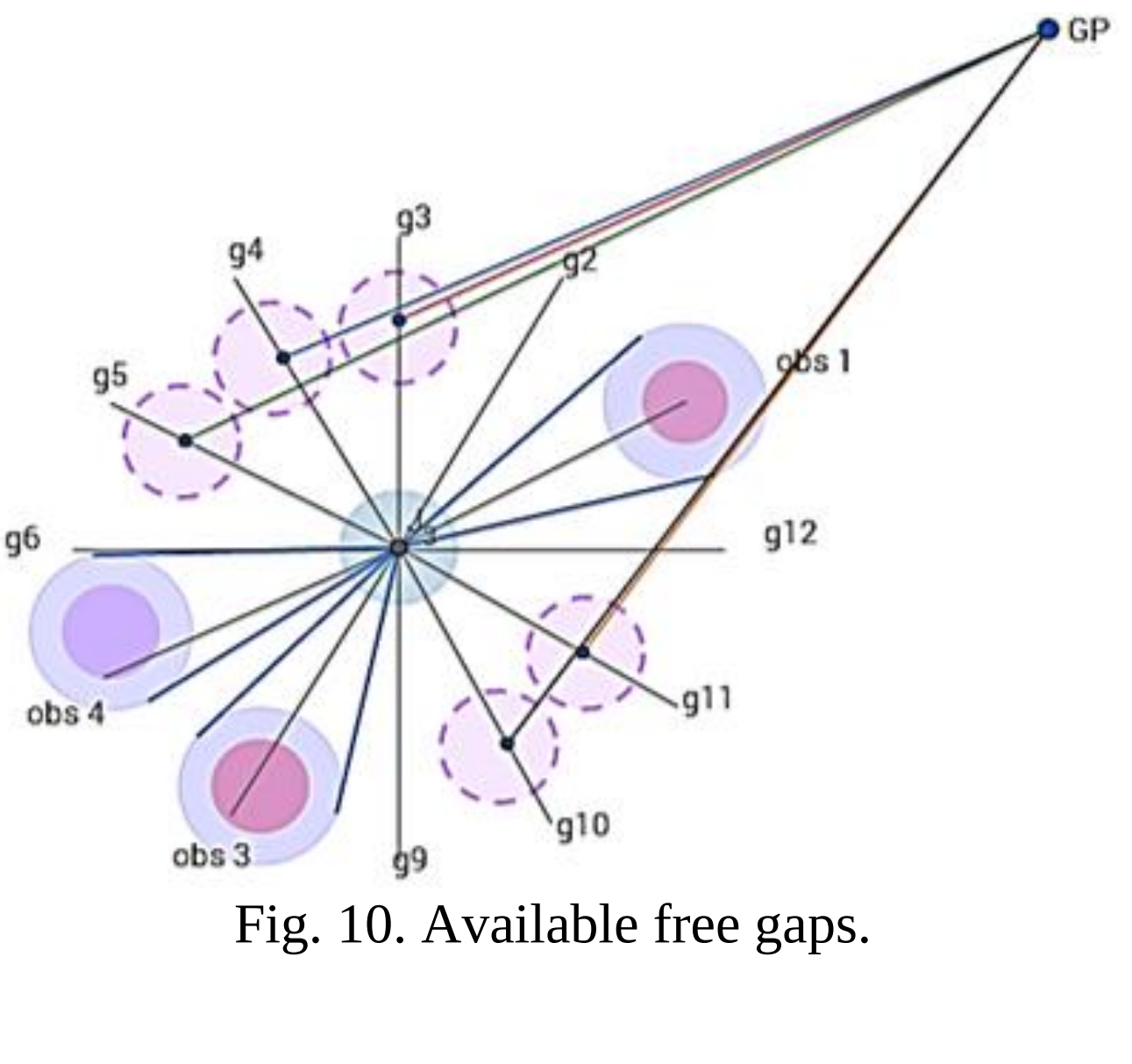

Fig. 10. Available free gaps.

---

**Algorithm 2:** Obstacle Avoidance (OA)

---

**Inputs:** Given $m$ is the number of obstacles existing within $SR$, $r_{MR}$, $MR_{Pos} = (x_{MR}, y_{MR})$, $r_{Obs}$, and $obs_i_{Pos} = (x_{obs_i}, y_{obs_i})$;

**Outputs:** Finding the best permissible mobile robot position $g\mu_{Pos}$ = $(x_{g\mu}, y_{g\mu})$;

1: **for** $i$=1 : $m$

2: $d_i = d(MR_{Pos}, obs_i_{Pos})$

3: **end for**

4: Calculate the shortest distance ($d_{sh}$),

5: $d_{sh}$ = min ( $d_1, \ldots, d_m$);

6: $Q$ = $r_{MR}$+$r_{Obs}$ +$SR$;

7: $\mu$ = index of the "zeros" in $Vg$;
$Mg$ = total number of available free gaps;

8: Calculate the new allowable mobile robot positions $gi_{Pos}$= $(x_{gi}, y_{gi})$; as follows

9: **for** $i$ = $\mu$: $Mg$

10: **if** ($d_{sh} < \frac{Q}{2}$) **then**

11: $x_{gi} = x_{MR} + 1.5 \times d_{sh} \times cos\,\psi_i$
$y_{gi} = y_{MR} + 1.5 \times d_{sh} \times sin\,\psi_i$

12: **Else**

13: $x_{gi} = x_{MR} + d_{sh} \times \cos\psi_i$
$y_{gi} = y_{MR} + d_{sh} \times \sin\psi_i$

14: **end if**

15: $GP_{Pos} = (x_{GP}, y_{GP})$;

16: $dgi = d(gi_{Pos}, GP_{Pos})$ *using* (1);

17: **end for**

18: The best allowable position $g\mu_{Pos}$= $(x_{g\mu}, y_{g\mu})$ is chosen by OA that has the smallest $d(g\mu_{Pos}, GP_{Pos})$;

Fig. 10 offers an illustrative example. There are five available gaps in $Vg$, labelled g3, g4, g5, g10, and g11. Assume that all the following positions and radii are in metres and that $MR_{Pos}$ = (3, 3), $obs1_{Pos}$ = (3.71, 3.41), $obs3_{Pos}$ = (2.31, 2.75), $obs4_{Pos}$ = (2.34, 2.76), $GP_{Pos}$= (10, 10), $SR$ = 0.8, and $r_{MR}$= $r_{Obs}$= 0.3. The Euclidean distance between the mobile robot and each obstacle can be calculated as $d(MR_{Pos}, obs1) = 0.82$, $d(MR_{Pos}, obs3) = 0.73$, and $d(MR_{Pos}, obs4) = 0.7$. Therefore, $d_{sh}$ = 0.7. It is obvious that all Euclidean distances between the mobile robot and each obstacle is larger than or equal to $\frac{Q}{2}$, where $Q = r_{MR}$+$r_{Obs}$ +$SR$. Specifically in this case, $Q$ = 0.3+0.3+0.8 = 1.4 and $d_{sh}$= $\frac{Q}{2}$ (line 13 in **Algorithm 2** is activated).Thus, the new allowable mobile robot positions at the available free gaps can be calculated according to **Algorithm 2** and are tabulated in Table 1.

TABLE 1: Calculations of the allowable mobile robot positions

| Gap index($\mu$) | Angle $\psi_i$ (deg) | Suggested position (x, y) | Distance to GP |
|---|---|---|---|
| 3 | 90° | (3, 3.7) | 9.4175 |
| 4 | 120° | (2.65, 3.6) | 9.7459 |
| 5 | 150° | (2.39, 3.35) | 10.1062 |
| 10 | 300° | (3.35, 2.39) | 10.1062 |
| 11 | 330° | (3.6, 2.75) | 9.6707 |

Thus, the new allowable mobile robot position $MR_New_{Pos} = (x_{MR_New}, y_{MR_New})$ will be at $g3_{Pos}$= $(x_{g3}, y_{g3})$ and is given as

$$x_{MR_New} = x_{g3} = 3 + 0.7 \times \cos(90) = 3$$
$$y_{MR_New} = y_{g3} = 3 + 0.7 \times \sin(90) = 3.7$$

The mobile robot will avoid the obstacles through gap g3, since this has shortest distance with GP, dg3= 9.4175 m, as in Fig.10.

*D. Proposed Complete Path Planning Algorithm*

In this subsection, the complete path planning algorithm for a mobile robot with an omnidirectional mobile robot is presented in static and dynamic environments. The first step for planning a path is to initialise the environment settings (*SP*, *GP*, $Obs_{Pos}$, *SR*) and the parameter settings of the proposed PSO-MFB optimization algorithm. The current position of the mobile robot ($MR_{Pos}$) is stored in a path vector called path.

The mobile robot continues gathering information about the surrounding environment via the deployed sensors to detect any obstacles while it navigates toward its *GP*. The overall process is presented in **Algorithm 3**. It should be emphasized that the process of the path planning problem in a dynamic environment is similar to that in static environment except that the movement of the dynamic obstacles in Equations (7-10) are considered.

**Algorithm 3:** Path Planning using hybrid PSO-MFB Algorithm- Static and Dynamic Environments

1: **Initialize:** $SP_{Pos}$, $GP_{Pos}$, $Obs_{Pos}$, $SR$=0.8, $r_{MR}$, $r_{Obs}$, PSO parameters: population size of $n$ particle, $r_1$, $r_2$, $c_1$, $c_2$, MFB parameters: population size of $m$ bats, frequency $f_i$, pulse rate $r_i$ and the loudness $A_i$;
2: $MR_{Pos} = SP_{Pos}$;
3: $path \leftarrow MR_{Pos}$;
4: **while** $(MR_{Pos} \neq \, , GP_{Pos})$
5:     **if** obstacles are dynamic **then**
6:         $Obs_{Pos} \leftarrow new\ Obs_{Pos}(Eq.(7-10))$
7:     **end if**
8: Collect data from virtual twelve sensors;
9:     **if** $d(MR_{Pos}, Obs_{Pos}) \leq Q$ **then**
10:         switch on [**Algorithm 2**];
11:     **Else**
12:         navigate toward GP [**Algorithm 1**];
13:     **end if**
14:     go to **3**
15: **end while**

## 6. Simulation Results

In simulation, three case studies were conducted. The first case study included simulations of path planning by the mobile robot in a static environment, while the second case study presents the simulation results in a dynamic environment, and the final case study contains a comparison with previous works. All experiments are achieved the following solutions after executing the algorithm ten times using MATLAB R2015a programming language. The MATLAB codes are run on a computer system with 2.76 GHz Core i7 CPU, and 4 G RAM.

### *A. Algorithmic Parameter Setting*

The simulation parameters for PSO-MFB algorithm are listed in Table 2.

TABLE 2: Parameter setting for Hybrid PSO-MFB

| Parameter | Value |
|---|---|
| No. of iteration | 5 |
| population size$_{PSO}$ | 5 |
| population size$_{MFBA}$ | 5 |
| $[c_1, c_2]$ | [2, 2] |
| $[f_{min}, f_{max}]$ | [0, 10] |
| σ | 0.3 |
| sensor range | 0.8 (m) |

The simulation parameters for the algorithms used in the comparisons section, presented in [33], are: The maximum cycle= 120, limit= 50, and 1 scout bee is generated each time, penalty $\delta$=100, the coefficients values $\alpha$=2 and $\beta$=0.5, population size is 6, where 3 are employed bees and the other 3 are onlooker bees; food source NS is 3 and $n_p$=3. While in [34] The GA used to simulate the cases had population size 50, crossover probability 0.85, mutation probability 0.15, size of each chromosome 16 bits (binary codification), maximum number of generations 100, the selection operator is roulette wheeling with elitist structure. For the bacteria colony algorithm the following parameters needed population size=50, number of chemotactic steps 15, maximum number of steps that a bacterium can swim in a turn =10, number of reproductions =4, number of elimination-dispersals events=2, elimination dispersal Probability=0.3, of the movement taken in one step =2.5.

### *B. Path Planning in a Static Environment*

In this case study, an environment with static obstacles is utilised to demonstrate the effectiveness of the proposed path planning algorithm for the mobile robot. The static environment consists of five static obstacles of different sizes. The starting point was $SP_{Pos}$ = (0, 0), the goal point was $GP_{Pos}$= (10, 10), and the radius of the mobile robot was $r_{MR}$ = 0.5 (m). The proposed **Algorithm 3** was applied and **Algorithm 1** was also applied as a point generation tool for the path planning algorithm in 3. The settings for this environment are listed in Table 3, and the optimized fitness function is defined as in (5).

TABLE 3: Static Environment Settings

| Obstacle no. | Radius($r_{OBS}$) | Position ($Obs_{Pos}$) |
|---|---|---|
| 1 | 0.5 | (2, 2.3) |
| 2 | 0.8 | (5, 4) |
| 3 | 1.2 | (8, 2) |
| 4 | 1 | (7.7, 7) |
| 5 | 0.7 | (3, 8.3) |

The best path (higher fitness of (6) with maximum smoothness and the shortest distance was obtained as equal to 14.7785 m as shown in Fig. 11, passing through points (2.7467, 1.5316), (2.9402, 1.9507), (3.8288, 4.7084), (6.6503, 8.1422), and (6.8160, 8.2288) with computation time (3.48) min. as shown in Table 4.

TABLE 4: Simulation result using MFB and Hybrid PSO-MFB algorithm

| Run no. | Path Length MFB | Fitness | Computation time (min) | Path Length (Hybrid PSO-MFB | Fitness | Computation time (min) |
|---|---|---|---|---|---|---|
| 1 | **14.793** | **0.067599** | **0.5529887** | 14.786 | 0.06763 | 3.425260 |
| 2 | 14.8112 | 0.067516 | 0.7384254 | 14.7953 | 0.06758 | 3.189191 |
| 3 | 14.793 | 0.067599 | 0.7050320 | 14.796 | 0.06758 | 3.704136 |
| 4 | 14.8112 | 0.067516 | 0.8844321 | 14.8083 | 0.06752 | 3.368553 |
| 5 | 14.8525 | 0.067328 | 1.2217404 | 14.7909 | 0.06760 | 3.316233 |
| 6 | 14.8028 | 0.067554 | 0.7457431 | **14.7785** | **0.06766** | **3.483598** |
| 7 | 14.8798 | 0.067205 | 0.8554462 | 14.7915 | 0.06760 | 3.720351 |
| 8 | 14.8051 | 0.067544 | 0.8571433 | 14.7876 | 0.06762 | 3.521768 |
| 9 | 14.793 | 0.067599 | 0.6038037 | 14.8023 | 0.06755 | 3.201012 |
| 10 | 14.8112 | 0.067516 | 0.7717397 | 14.7921 | 0.06760 | 3.247233 |

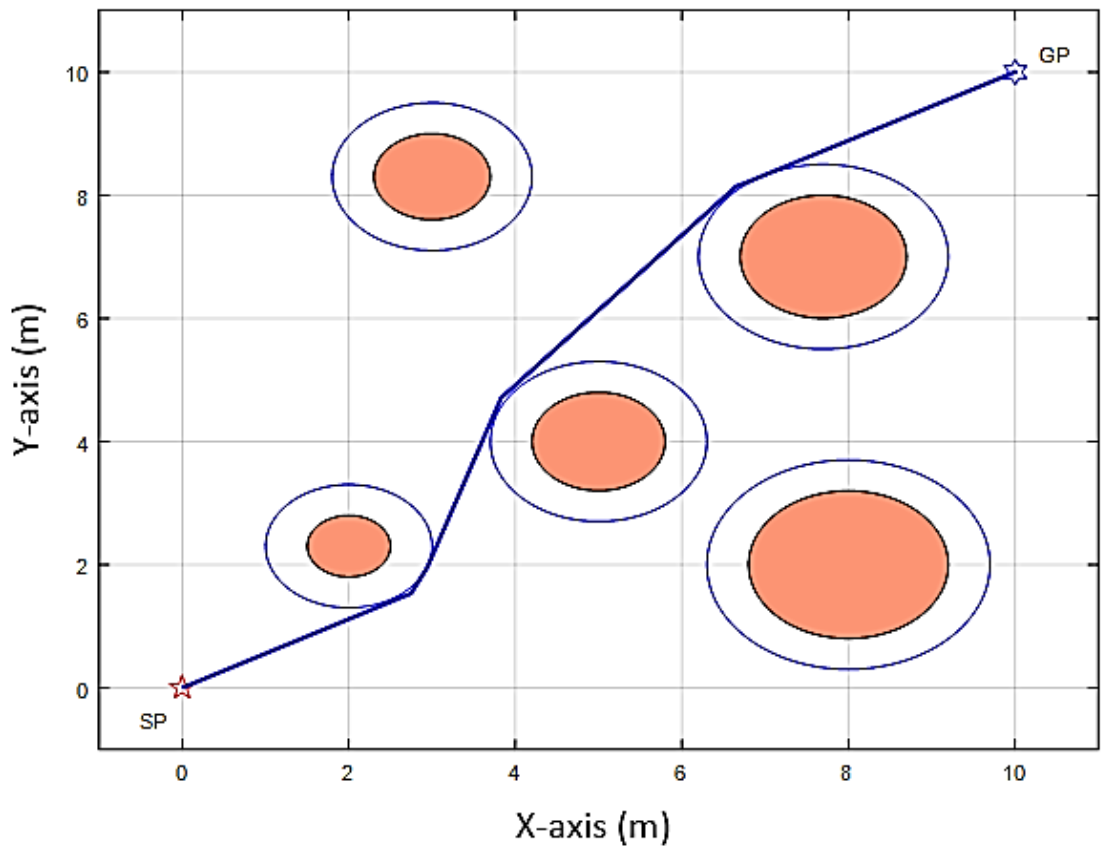

Fig. 11. The best path achieved using Algorithm 3.

*C. Path Planning in a Dynamic Environment*

In the second case study, the proposed path planning **Algorithm 3** was tested under a dynamic environment consisting of six dynamic obstacles; the starting point was $SP_{Pos}$ = (1, 1), the goal point was $GP_{Pos}$= (10, 10), and the radius of the mobile robot was $r_{MR}$=

0.3 m. In this environment, some of the obstacles moved linearly and others moved in a circular trajectory. The positions, velocities, and directions of the dynamic obstacles are listed in Tables 5 and 6.

TABLE 5: Settings for the linear moving obstacles

| Obstacle no. | Centre | Radius($r_{OBS}$) | $v_{obs}$ (m/s) | $\varphi_{obs}$ (deg) |
|---|---|---|---|---|
| 1 | (7.5,2.1) | 0.3 | 0.16 | 70° |
| 2 | (5.1,8.3) | 0.3 | 0.13 | 0° |

TABLE 6: Settings for the circular moving obstacles

| Obstacle no. | Initial Position | Circle Centre $(x_c, y_c)$ | Circle Radius $(r_c)$ |
|---|---|---|---|
| 3 | (6,5) | (5, 5) | 1 |
| 4 | (4,5) | (5, 5) | 1 |
| 5 | (5,7.5) | (5, 5) | 2.5 |
| 6 | (5,2.5) | (5, 5) | 2.5 |

The mobile robot navigates from its *SP* toward its *GP* using the proposed **Algorithm 3** as shown in Fig.12 (a) until it encounters an obstacle within its sensing region as depicted in Fig. 12 (b-e). At that time, the mobile robot triggers **Algorithm 2** to avoid the obstacles and changes its original path, using the proposed new collision-free path toward *GP*. The mobile robot continues its motion using **Algorithm 3** until it reaches GP. The best collision-free path obtained was 13.6696 m with computation time (4.162) sec, as shown in Fig. 12(f).

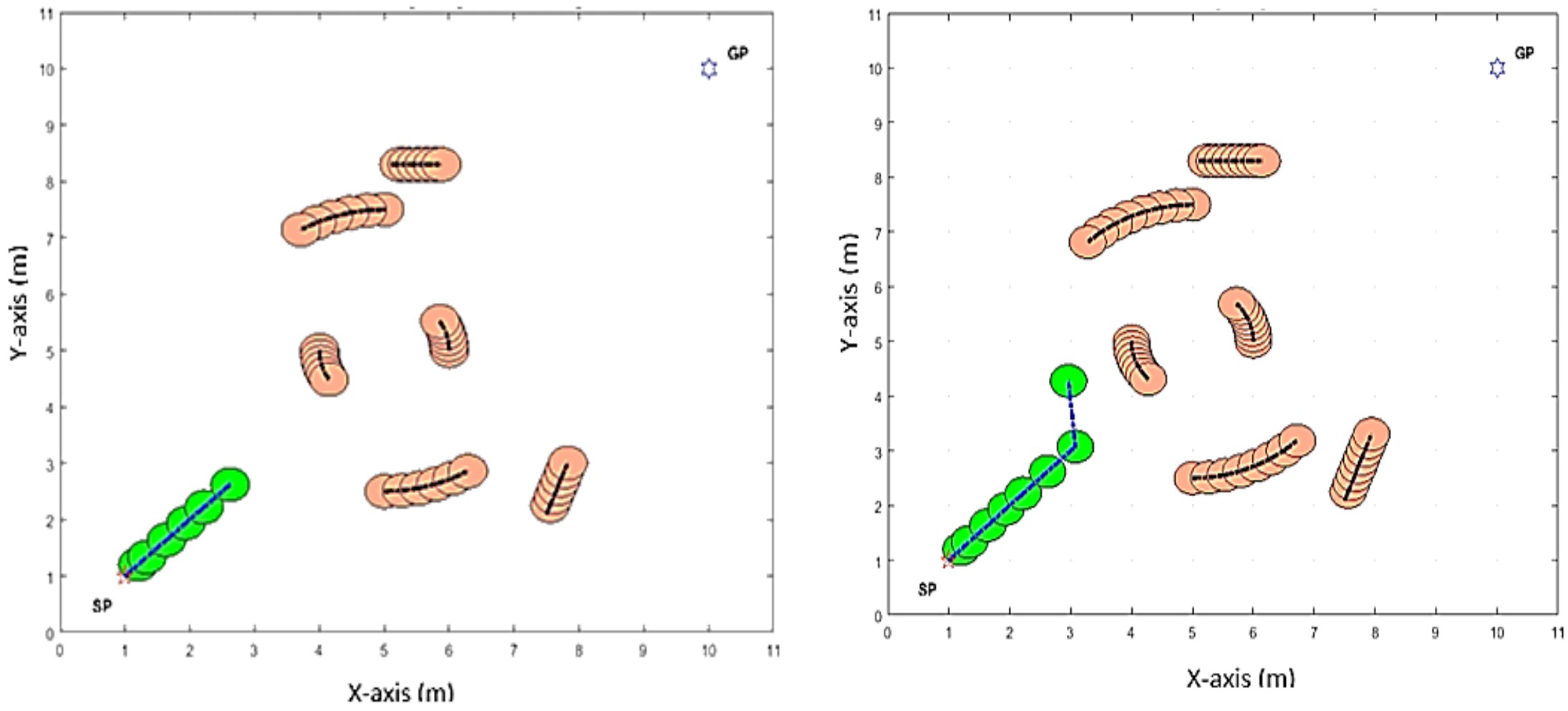

(a) (b)

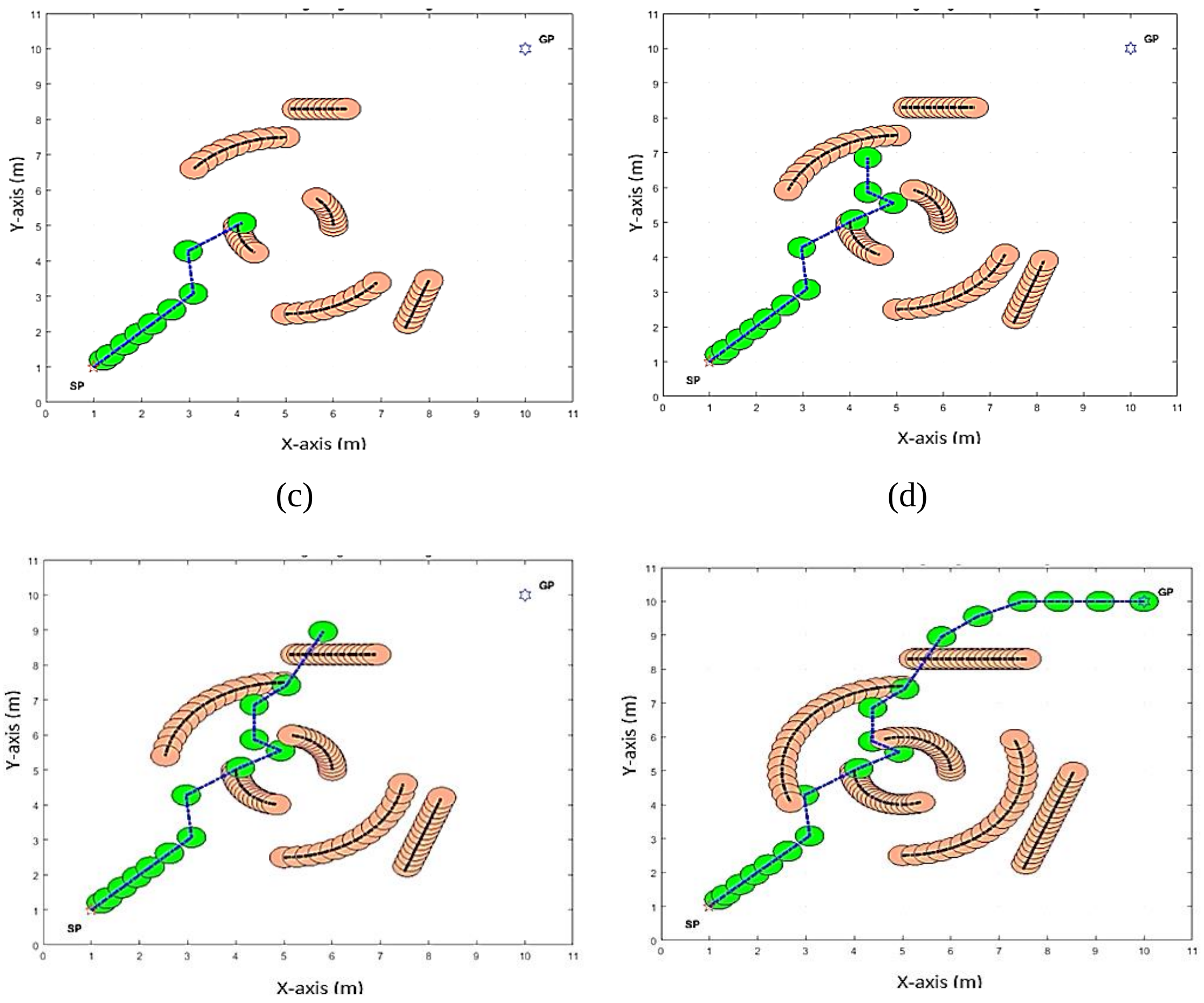

(c) (d)

(e) (f)

Fig. 12. The best path achieved with Algorithm 3 for the above dynamic environment.

### *D. Comparison with Other Path Planning Algorithms*

In this subsection, the performance of the proposed path planning algorithm using a hybrid PSO-MFB algorithm is compared with the works of [33–35]. The first case study involved an environment also used in these works, which consists of four static obstacles. The optimization techniques used to obtain the best path were the Direct Artificial Bee Colony (DABC) and Minimum Angle Artificial Bee Colony (MAABC) algorithms in [33] while the work in [34] included GA and Bacterial Colony (BC) algorithms. The proposed Hybrid PSO-MFB algorithm was applied in the same environment as shown in Fig. 13.

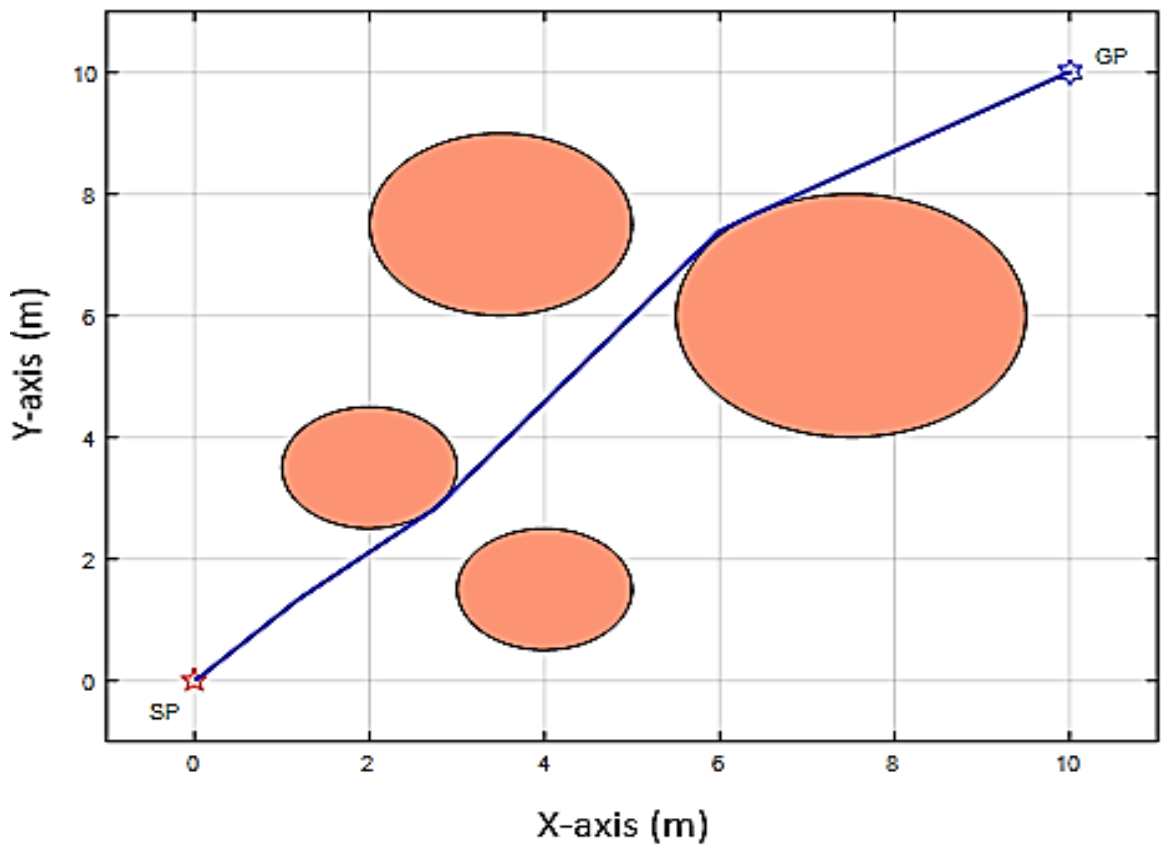

Fig. 13. The best path achieved using Algorithm 3.

The best path obtained using the proposed Hybrid PSO-MFB Algorithm was 14.3255 m, compared with 14.3625 m using DABC, 14.3371 m using MAABC, 14.5095 m using GA, and 14.3802 m using BC. The comparison with these works is illustrated in Table 7. It is worth mentioning that the environment size in [35] was 100m x 100m, and that this environment was scaled to 10m x 10m by dividing the results of GA and BC algorithms by a factor of 10 to make a fair comparison between the proposed PSO-MFB algorithm and the GA and BC ones.

TABLE 7: Comparison Results for the first case study with ten experiments

| Optimization Technique | Max. Fitness | Min. Fitness | Mean Fitness |
|---|---|---|---|
| Hybrid PSO-MFB | **0.06980** | **0.06968** | **0.06977** |
| DABC | 0.06962 | 0.06859 | 0.06934 |
| MAABC | 0.06974 | 0.06865 | 0.06934 |
| GA | 0.06892 | 0.06448 | 0.06778 |
| BC | 0.06954 | 0.06908 | 0.06937 |

Based on the above results, it is obvious that the Hybrid PSO-MFB algorithm outperforms the other optimization techniques listed in Table 7. The proposed hybrid PSO-MFB algorithm provides the best path, which can be verified by the values of the mean, minimum, and maximum fitness values.

The environment in [35], which involved six static obstacles, was used in the second case study for comparison. The optimization technique used to obtain the shortest path was a standard ABC in [35], and DABC and MAABC in [33]. The proposed Hybrid PSO-MFB algorithm was applied to the same environment, as shown in Fig. 14.

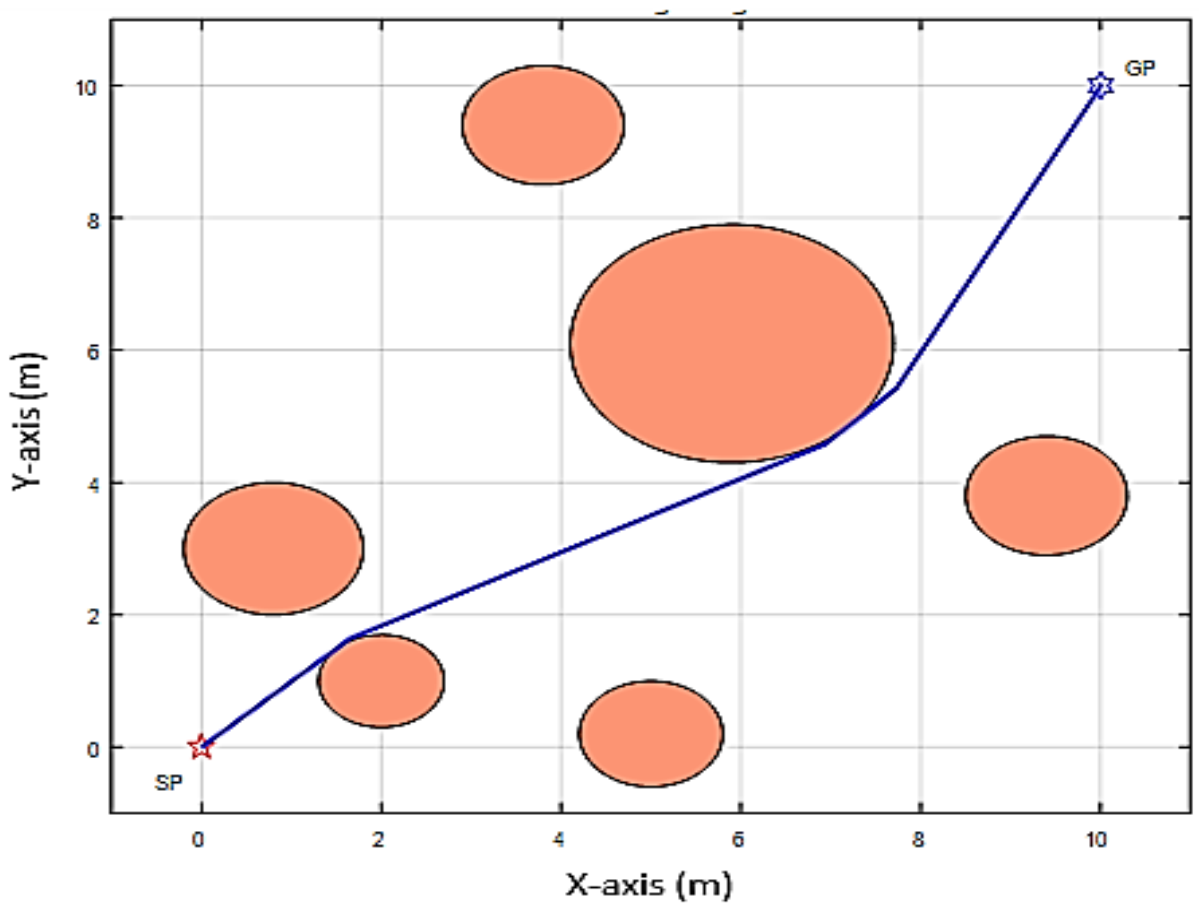

Fig. 14. The best path obtained using Algorithm 3.

The best path obtained using the proposed Hybrid PSO-MFB algorithm was 14.6384 m, compared to 14.7422 m and 14.7163 m using the DABC, MAABC algorithms, respectively. The best path using the ABC algorithm in [35] was 14.8821 m (after scaling by 10). The comparison results show that the proposed Hybrid PSO-MFB Algorithm outperforms ABC, DABC, and MAABC optimization algorithms in terms of finding the shortest distance. Finally, even though the improvement in the shortest path using the proposed Hybrid PSO-MFB algorithm is slight as compared with some path planning algorithms, in environments with large distances, the slight improvements in the shortest path become noticeable.

## 7. Discussion

Based on the simulation results it is obvious that the proposed Hybrid PSO-MFB algorithm outperforms other optimization techniques in terms of the shortest distance (i.e., it has higher fitness value). More specifically, the hybrid PSO-MFB outperforms the MFB in terms of path length because each particle in PSO calls for a complete MFB algorithm with dynamic parameters ($\alpha$ &$\gamma$) rather than static ones. This leads to more adaptation of the MFB algorithm to the loudness $A_i(t)$ and the rate of the pulse emission $r_i(t)$, which increases the potential of the MFB algorithm to find better solutions. However, this success is on account of the computation time of the algorithm, whereas with Hybrid PSO-MFB algorithm it is higher than that of the MFB one. Fortunately, this problem can be overcome easily with the advancement in the H/W technology and the development of small-sized and very high-speed microcontroller units and FPGA boards. Concerning the comparisons with the already existing path planning methods in the literature, more improvement can be achieved using the proposed PSO-MFB algorithm, but, on account of the mobile robot safety (i.e., mobile robot gets too close to one or more of the obstacles).

## 8. Conclusions

This paper proposed a path planning algorithm for mobile robots using a Hybrid PSO-MFB swarm optimization algorithm integrated with LS and ODA strategies. The size of the mobile robot was taken into account by enlarging the size of the obstacles in the free-space environment. The algorithm was tested in static and dynamic environments with different scenarios to minimise a multi-objective measure of path length and minimum angles. In the context of the simulation results, it can be concluded that the proposed hybrid PSO-MFB algorithm proved its efficacy in avoiding static and dynamic obstacles in a simple manner and reduced time. The simulation results demonstrated that the proposed path planning algorithm offers significant advances over current state-of-the-art options. In future work, consideration of the H/W implementation of the proposed Hybrid PSO-MFB algorithm-based path planning on real omnidirectional mobile platform will be an interesting task. Another possible direction which might be conducted in the future is the implementation of the proposed path planning algorithm in an intricate cluttered dynamic environment with moving target.

## References

[1] B.K. Patle, G. Babu L, A. Pandey, D.R.K. Parhi, A. Jagadeesh, A review: On path planning strategies for navigation of mobile robot, Def. Technol. 15 (2019) 582–606. https://doi.org/10.1016/j.dt.2019.04.011.

[2] H. Cheon, B.K. Kim, Online Bidirectional Trajectory Planning for Mobile Robots in State-Time Space, IEEE Trans. Ind. Electron. 66 (2019) 4555–4565. https://doi.org/10.1109/TIE.2018.2866039.

[3] S. Hosseininejad, C. Dadkhah, Mobile robot path planning in dynamic environment based on cuckoo optimization algorithm, Int. J. Adv. Robot. Syst. 16 (2019) 172988141983957. https://doi.org/10.1177/1729881419839575.

[4] J.M. Keil, DECOMPOSING A POLYGON INTO SIMPLER COMPONENTS., SIAM J. Comput. 14 (1985) 799–817. https://doi.org/10.1137/0214056.

[5] C. Zhong, S. Liu, B. Zhang, Q. Lu, J. Wang, Q. Wu, F. Gao, A Fast On-line Global Path Planning Algorithm Based on Regionalized Roadmap for Robot Navigation, IFAC-PapersOnLine. 50 (2017) 319–324. https://doi.org/10.1016/j.ifacol.2017.08.053.

[6] U. Orozco-Rosas, O. Montiel, R. Sepúlveda, Mobile robot path planning using membrane evolutionary artificial potential field, Appl. Soft Comput. J. 77 (2019) 236–251. https://doi.org/10.1016/j.asoc.2019.01.036.

[7] T.T. Mac, C. Copot, D.T. Tran, R. De Keyser, Heuristic approaches in robot path planning: A survey, Rob. Auton. Syst. 86 (2016) 13–28. https://doi.org/10.1016/j.robot.2016.08.001.

[8] R. Rashid, N. Perumal, I. Elamvazuthi, M.K. Tageldeen, M.K.A.A. Khan, S. Parasuraman, Mobile robot path planning using Ant Colony Optimization, in: 2016 2nd IEEE Int. Symp. Robot. Manuf. Autom. ROMA 2016, Institute of Electrical and Electronics Engineers Inc., 2017.

https://doi.org/10.1109/ROMA.2016.7847836.

[9] I.K. Ibraheem, F.H. Ajeil, Path Planning of an autonomous Mobile Robot using Swarm Based Optimization Techniques, Al-Khwarizmi Eng. J. 12 (2017) 12–25. https://doi.org/10.22153/kej.2016.08.002.

[10] J. Ou, M. Wang, Path planning for omnidirectional wheeled mobile robot by improved ant colony optimization, in: Chinese Control Conf. CCC, IEEE Computer Society, 2019: pp. 2668–2673. https://doi.org/10.23919/ChiCC.2019.8866228.

[11] G. Wang, L. Guo, H. Duan, L. Liu, H. Wang, A bat algorithm with mutation for UCAV path planning, Sci. World J. 2012 (2012). https://doi.org/10.1100/2012/418946.

[12] H.S. Dewang, P.K. Mohanty, S. Kundu, A Robust Path Planning for Mobile Robot Using Smart Particle Swarm Optimization, in: Procedia Comput. Sci., Elsevier B.V., 2018: pp. 290–297. https://doi.org/10.1016/j.procs.2018.07.036.

[13] W. Wang, M. Cao, S. Ma, C. Ren, X. Zhu, H. Lu, Multi-robot odor source search based on Cuckoo search algorithm in ventilated indoor environment, in: Proc. World Congr. Intell. Control Autom., Institute of Electrical and Electronics Engineers Inc., 2016: pp. 1496–1501. https://doi.org/10.1109/WCICA.2016.7578817.

[14] M.A. Hossain, I. Ferdous, Autonomous robot path planning in dynamic environment using a new optimization technique inspired by bacterial foraging technique, Rob. Auton. Syst. 64 (2015) 137–141. https://doi.org/10.1016/j.robot.2014.07.002.

[15] P.K. Das, S.K. Pradhan, S.N. Patro, B.K. Balabantaray, Artificial immune system based path planning of mobile robot, Stud. Comput. Intell. 395 (2012) 195–207. https://doi.org/10.1007/978-3-642-25507-6_17.

[16] T.K. Dao, T.S. Pan, J.S. Pan, A multi-objective optimal mobile robot path planning based on whale optimization algorithm, in: Int. Conf. Signal Process. Proceedings, ICSP, Institute of Electrical and Electronics Engineers Inc., 2016: pp. 337–342. https://doi.org/10.1109/ICSP.2016.7877851.

[17] C. Lamini, S. Benhlima, A. Elbekri, Genetic algorithm based approach for autonomous mobile robot path planning, in: Procedia Comput. Sci., Elsevier B.V., 2018: pp. 180–189. https://doi.org/10.1016/j.procs.2018.01.113.

[18] U.A. Syed, F. Kunwar, M. Iqbal, Guided Autowave Pulse Coupled Neural Network (GAPCNN) based real time path planning and an obstacle avoidance scheme for mobile robots, Rob. Auton. Syst. 62 (2014) 474–486. https://doi.org/10.1016/j.robot.2013.12.004.

[19] D. Davis, P. Supriya, Implementation of fuzzy-based robotic path planning, Adv. Intell. Syst. Comput. 380 (2016) 375–383. https://doi.org/10.1007/978-81-322-2523-2_36.

[20] A. Pandey, D.R. Parhi, Optimum path planning of mobile robot in unknown static

and dynamic environments using Fuzzy-Wind Driven Optimization algorithm, Def. Technol. 13 (2017) 47–58. https://doi.org/10.1016/j.dt.2017.01.001.

[21] D.R. Parhi, P.K. Mohanty, IWO-based adaptive neuro-fuzzy controller for mobile robot navigation in cluttered environments, Int. J. Adv. Manuf. Technol. 83 (2016) 1607–1625. https://doi.org/10.1007/s00170-015-7512-5.

[22] T.A.V. Teatro, J.M. Eklund, R. Milman, Nonlinear model predictive control for omnidirectional robot motion planning and tracking with avoidance of moving obstacles, Can. J. Electr. Comput. Eng. 37 (2014) 151–156. https://doi.org/10.1109/CJECE.2014.2328973.

[23] O. Castillo, P. Melin, F. Valdez, R. Martínez-Marroquín, Optimization of Fuzzy Logic Controllers for Robotic Autonomous Systems with PSO and ACO, in: 2011: pp. 389–417. https://doi.org/10.1007/978-3-642-17390-5_17.

[24] W.J. Chen, B.G. Jhong, M.Y. Chen, Design of Path Planning and Obstacle Avoidance for a Wheeled Mobile Robot, Int. J. Fuzzy Syst. 18 (2016) 1080–1091. https://doi.org/10.1007/s40815-016-0224-7.

[25] A. Bakdi, A. Hentout, H. Boutami, A. Maoudj, O. Hachour, B. Bouzouia, Optimal path planning and execution for mobile robots using genetic algorithm and adaptive fuzzy-logic control, Rob. Auton. Syst. 89 (2017) 95–109. https://doi.org/10.1016/j.robot.2016.12.008.

[26] M. Emam, A. Fakharian, Path following of an omni-directional four-wheeled mobile robot, in: 2016 Artif. Intell. Robot. IRANOPEN 2016, Institute of Electrical and Electronics Engineers Inc., 2016: pp. 36–41. https://doi.org/10.1109/RIOS.2016.7529487.

[27] H. Kim, B.K. Kim, Minimum-energy cornering trajectory planning with self-rotation for three-wheeled omni-directional mobile robots, Int. J. Control. Autom. Syst. 15 (2017) 1857–1866. https://doi.org/10.1007/s12555-016-0111-x.

[28] X. Wang, Y. Shi, D. Ding, X. Gu, Double global optimum genetic algorithm-particle swarm optimization-based welding robot path planning, Eng. Optim. 48 (2016) 299–316. https://doi.org/10.1080/0305215X.2015.1005084.

[29] J.H. Zhang, Y. Zhang, Y. Zhou, Path planning of mobile robot based on hybrid multi-objective bare bones particle swarm optimization with differential evolution, IEEE Access. 6 (2018) 44542–44555. https://doi.org/10.1109/ACCESS.2018.2864188.

[30] M. Saraswathi, G.B. Murali, B.B.V.L. Deepak, Optimal Path Planning of Mobile Robot Using Hybrid Cuckoo Search-Bat Algorithm, in: Procedia Comput. Sci., Elsevier B.V., 2018: pp. 510–517. https://doi.org/10.1016/j.procs.2018.07.064.

[31] R.C. Eberhart, J. Kennedy, R.C. Eberhart, Particle swarm optimization, Proc. IEEE Int. Conf. Neural Networks IV, Pages. 4 (1995) 1942–1948. https://doi.org/10.1109/ICNN.1995.488968.

[32] X.S. Yang, A new metaheuristic Bat-inspired Algorithm, in: Stud. Comput. Intell., 2010: pp. 65–74. https://doi.org/10.1007/978-3-642-12538-6_6.

[33] N. HadiAbbas, F. Mahdi Ali, Path Planning of an Autonomous Mobile Robot using Directed Artificial Bee Colony Algorithm, Int. J. Comput. Appl. 96 (2014) 11–16. https://doi.org/10.5120/16836-6681.

[34] C.A. Sierakowski, L. Dos, S. Coelho, STUDY OF TWO SWARM INTELLIGENCE TECHNIQUES FOR PATH PLANNING OF MOBILE ROBOTS, 2005.

[35] J.-H. Lin, L.-R. Huang, Chaotic Bee Swarm Optimization Algorithm for Path Planning of Mobile Robots, n.d.